\pdfoutput=1

\documentclass[11pt]{article}

\usepackage[]{ACL2023}

\usepackage{times}
\usepackage{latexsym}
\usepackage{amssymb}
\usepackage{amsmath}
\usepackage{multirow}
\usepackage{booktabs}  
\usepackage{makecell}  
\usepackage[T1]{fontenc}
\usepackage{graphicx}
\usepackage{xltabular}
\usepackage{stfloats}
\usepackage{rotating}    

\usepackage[utf8]{inputenc}  
\usepackage{microtype}
\usepackage{natbib}

\usepackage{inconsolata}
\usepackage{longtable}
\usepackage{diagbox}   
\usepackage{adjustbox} 
\title{Automated CAD Modeling Sequence Generation from Text Descriptions via Transformer-Based Large Language Models}

\author{\textbf{Jianxing Liao}\textsuperscript{1}\thanks{\quad These authors are equal first authors} , 
\textbf{Junyan Xu}\textsuperscript{1}\footnotemark[1] , 
\textbf{Yatao Sun}\textsuperscript{1} , 
\textbf{Maowen Tang}\textsuperscript{1} , 
\textbf{Sicheng He}\textsuperscript{1} ,\\
\textbf{Jingxian Liao}\textsuperscript{1},
\textbf{Shui Yu}\textsuperscript{1}\thanks{\quad Corresponding authors} , 
\textbf{Yun Li}\textsuperscript{1}\footnotemark[2]  ,
\textbf{Xiaohong Guan}\textsuperscript{2}
\\
\textsuperscript{1}Shenzhen Institute for Advanced Study, \\University of Electronic Science and Technology of China,\\
Shenzhen, 518000, China \\
\textsuperscript{2}i4AI Ltd, London WCIN3AX, United Kingdom \\
\footnotemark[2] Shui.Yu@i4AI.org, Yun.Li@ieee.org}

\begin{document}
\maketitle
\begin{abstract}
Designing complex computer-aided design (CAD) models is often time-consuming due to challenges such as computational inefficiency and the difficulty of generating precise models. We propose a novel language-guided framework for industrial design automation to address these issues, integrating large language models (LLMs) with computer-automated design (CAutoD).Through this framework, CAD models are automatically generated from parameters and appearance descriptions, supporting the automation of design tasks during the detailed CAD design phase. Our approach introduces three key innovations: (1) a semi-automated data annotation pipeline that leverages LLMs and vision-language large models (VLLMs) to generate high-quality parameters and appearance descriptions; (2) a Transformer-based CAD generator (TCADGen) that predicts modeling sequences via dual-channel feature aggregation; (3) an enhanced CAD modeling generation model, called CADLLM, that is designed to refine the generated sequences by incorporating the confidence scores from TCADGen. Experimental results demonstrate that the proposed approach outperforms traditional methods in both accuracy and efficiency, providing a powerful tool for automating industrial workflows and generating complex CAD models from textual prompts. The code is available at \url{https://jianxliao.github.io/cadllm-page/}
\end{abstract}

\section{Introduction}

Computer-aided design (CAD) is important in industrial design and additive manufacturing. It is a key tool from the early stages of design to prototype production~\cite{intro1-marchesi2021chairside, intro2-rapp2021mlcad}. Computer-Automated Design (CAutoD) represents an evolution of the traditional CAD by integrating automation and machine learning algorithms to assist in the design process~\cite{li2022should}. With the fast growth of AI, especially large language models (LLMs), combining CAD and AI is becoming a major factor in improving design speed and creativity~\cite{intro3-guo2022complexgen, intro4-dai2024fabric}. This development speeds up the design process and increases the possible uses of CAD systems, making them more intelligent and personalized. In recent years, more research focuses on combining AI with CAD~\cite{intro6-xu2024cad, intro8-wu2024cadvlm}. Methods such as generating parametric CAD models from the textual description and improving automated design tools with AI are proposed~\cite{Text2cad-yavartanoo2024text2cad}. These studies show that AI in CAD can speed up design and make CAD tools more flexible and smart. The integration of AI is pushing CAD towards CAutoD, representing a shift from the traditional, computer-aided design to computer-automated design processes.

Most current research on combining LLMs with CAD focuses on the data synthesis ability of these models, such as generating design information by processing large datasets. However, design data synthesized by LLMs often lacks sufficient accuracy~\cite{dataset1-fan2025parametric, deepcad-wu2021deepcad}. Without parameter review, converting this data into high-quality, editable CAD models is difficult, especially in industrial design, where precision is critical. While current research emphasizes the ability of LLMs to understand complex problems, their application in complex CAD tasks still faces several challenges~\cite{llmre1-plaat2024reasoning}. First, LLMs are computationally expensive and inefficient~\cite{llmre2-luo2023reasoning}. Second, it is difficult for designers to guide these LLMs to generate reasonable CAD models using simple language or visual descriptions, limiting their practical use in CAD design. Despite these challenges, LLMs still have significant potential in CAD design and warrant further research. One promising solution is to combine finely tuned small generative models (with fewer parameters) with large models (with massive parameters)~\cite{small-xu2023small}. The small generative models can provide more accurate guidance for specific tasks, compensating for the limitations of the large models in detailed reasoning and thus improving overall accuracy~\cite{small2-chen2024role}. This approach can enhance the design synthesis capabilities of large models and reduce the uncertainties that often arise in the reasoning process~\cite{small3-fang2024improve}. Using small models as auxiliary modules in CAD design can improve the accuracy and efficiency of design, offering a more reliable solution for complex tasks.

\begin{figure*}[t]
\centering
\includegraphics[width=0.85\textwidth]{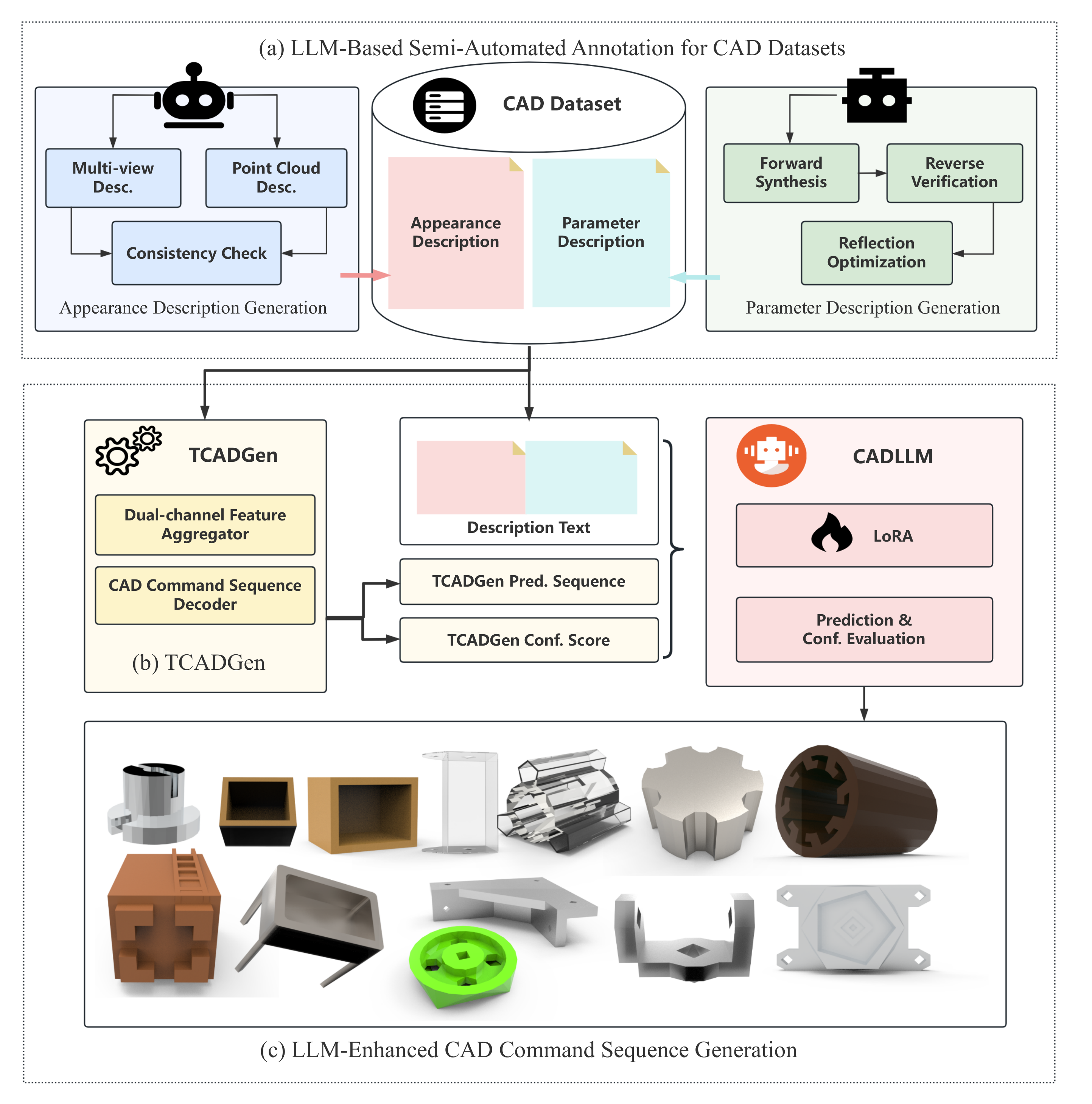}
  \caption{The overall framework of automated CAD modeling from text descriptions, leveraging Transformer-based sequence generation and LLM-driven refinement. The innovations proposed in our work are highlighted in (a), (b), and (c). }
  \label{fig:framework}
\end{figure*}

In this paper, we propose a novel framework, as shown in Fig.~\ref{fig:framework}, for generating high-quality CAD modeling sequences, where LLMs generate CAD operation commands and the corresponding parameters to facilitate CAD modeling. Specifically, we focus on the detailed design phase, where precise parameter and appearance descriptions are transformed into CAD models without human intervention. Through this automation, our work represents a significant step toward fully automating the CAD design process, enhancing both speed and accuracy in industrial applications, and supporting the vision of CAutoD. Our approach has three key innovations: 
\begin{itemize}
    \item We propose a semi-automated annotation pipeline with LLMs and VLLMs, enabling high-quality CAD description generation through automated validation and verification.
    \item We present TCADGen, a Transformer-based dual-channel architecture for transforming descriptive annotations into CAD commands, achieving improved sequence accuracy.
    \item We introduce a CAD-specific LLM enhancement framework, demonstrating its effectiveness in refining command types and parameter values for industrial applications.
\end{itemize}

\section{Related Works}

Extending the traditional
CAD, CAutoD has witnessed significant advancements in recent years across various industries, including ship design, architectural design, and mechanical engineering~\cite{ang2016smart,bye2017intelligent}. 
These advancements are largely driven by progress in dataset~\cite{deepcad-wu2021deepcad, Text2cad-yavartanoo2024text2cad, dataset1-fan2025parametric}, sequence generation~\cite{Text2cad-yavartanoo2024text2cad}, and large language models~\cite{intro2-rapp2021mlcad}. ModelNet is the first large dataset of 3D CAD mesh models ~\cite{Modelnet-vishwanath2009modelnet}. However, it does not include modeling process data. The Fusion 360 Gallery dataset later adds modeling history and assembly data~\cite{fusion360-willis2021fusion}. The Mechanical Components Benchmark then provides complete modeling histories for mechanical parts~\cite{mechanical-components-benchmark-kim2020large}. These datasets focus on geometric features but lack natural language descriptions of CAD models. Deep learning creates new ways to approach CAD modeling. DeepCAD uses a Transformer model to generate CAD sequences~\cite{deepcad-wu2021deepcad}. It breaks down CAD modeling into sketches and extrusion steps. However, it generates sketches before extrusion parameters, often creating disconnected sequences. It also supports only basic modeling operations. Text2CAD proposes a different approach that generates CAD sequences from text description and visual features~\cite{Text2cad-yavartanoo2024text2cad}. However, they do not consider that large models excel in generative and reasoning abilities to assist in design.

LLMs lead to new CAD modeling methods. BlenderLLM uses LLMs to generate CAD sequences through self-improvement~\cite{blenderllm-du2024blenderllm}. It creates the BlendNet dataset and CADBench for evaluation. However, its sequences lack geometric constraints, which affect model accuracy. Query2CAD uses LLMs to generate CAD commands and improves designs through iterations~\cite{query2cad-badagabettu2024query2cad}. However, it often misses steps when handling complex tasks. CADCodeVerify combines vision and language models to generate and improve CAD code~\cite{CADCodeVerify-alrashedy2024generating}. However, the multiple rounds of visual checking and code fixing needed make it slow to use. The AutoForma framework~\cite{liao2024autoforma}
, which leverages a large language model-based multi-agent system for the automated generation of CAD models. However, the method is costly and suffers from low generation efficiency.
\section{Methodology}
\subsection{Problem Definition}
Given a CAD model's appearance description text $T_\text{appear}\in \mathcal{V}_\text{appear}$ and its corresponding parameter modeling description text $T_\text{param} \in \mathcal{V}_\text{param}$, where $\mathcal{V}_\text{appear}$ and $\mathcal{V}_\text{param}$ represent the vocabulary for appearance descriptions and parameter descriptions respectively. Our goal is to generate a complete CAD Command Sequence(CCS)\footnote{An introduction to CCS can be found in Appendix~\ref{sec:CCS_introduction}}.  $\mathbf{M} = \{(c_1, \mathbf{p}_1), (c_2, \mathbf{p}_2), ..., (c_N, \mathbf{p}_N)\}$, where $c_i \in \mathcal{C}$ represents the type of the i-th modeling command, $\mathbf{p}_i \in \mathbb{R}^d$ represents the corresponding parameter vector in d-dimensional vector space, and $\mathcal{C}$ is the predefined set of CAD commands. This task can be divided into two stages: 

\textbf{TCADGen Sequence Generation:} We define the TCADGen sequence generation problem as a mapping from appearance and parameter descriptions to CAD command sequences and their confidence scores: $f_\text{TCADGen}(T_\text{appear}, T_\text{param}) = (\mathbf{M}, \mathbf{S})$, where $T_\text{appear}$ is the appearance description text, $T_\text{param}$ is the parameter description text, $\mathbf{M}$ is the predicted CAD Command Sequence (CCS), $\mathbf{S} = {s_1, s_2, ..., s_N}$ represents the confidence scores for each command and its parameters, $s_i = (s_i^\text{cmd}, s_i^\text{args})$ represents the confidence scores for command type and parameters respectively, $N$ is the total number of commands in the sequence, $s_i^\text{cmd}$ represents the confidence score for the $i$-th command type, and $s_i^\text{args}$ represents the confidence score for the parameters of the $i$-th command.

\textbf{CADLLM CCS enhancement:} We formulate the CADLLM CCS enhancement stage as: $ f_\text{CADLLM} (T_\text{appear}, T_\text{param}, \mathbf{M}, \mathbf{S}) = \mathbf{M}^*,$
where $\mathbf{M}^*$ is the final optimized modeling sequence, $f_\text{{CADLLM}}$ represents the industrial design domain large model CADLLM. The optimization objective is formulated as follows:
\begin{equation}
\begin{aligned}
& \max_{\mathbf{M}^*} \quad p(\mathbf{M}^*|T_\text{appear}, T_\text{param}) \\
& \text{s.t.} \quad \forall (c_i, \mathbf{p}_i) \in \mathbf{M}^*, c_i \in \mathcal{C}, \mathbf{p}_i \in \mathbb{R}^d,
\end{aligned}
\end{equation}
where $\mathbf{M}^{*}$ is constrained to be a valid CAD modeling sequence, $c_i$ represents the $i$-th command type, $\mathbf{p}_i$ represents the parameters for the $i$-th command, $\mathcal{C}$ is the set of all valid CAD commands, $d$ is the dimension of the parameter space, and $p(\mathbf{M}^|T_\text{appear}, T_\text{param})$ represents the probability of generating sequence $\mathbf{M}^*$ given the input descriptions.

The proposed framework aims to generate high-quality CAD modeling sequences through this two-stage process, where TCADGen generates the initial modeling sequences and the corresponding confidence assessment, followed by the optimization and improvement of the sequences in CADLLM. 
\subsection{Proposed Method}
\subsubsection{LLM-Based Semi-Automated Annotation for CAD Datasets}

\begin{figure*}
    \centering
    \includegraphics[width=0.85\linewidth]{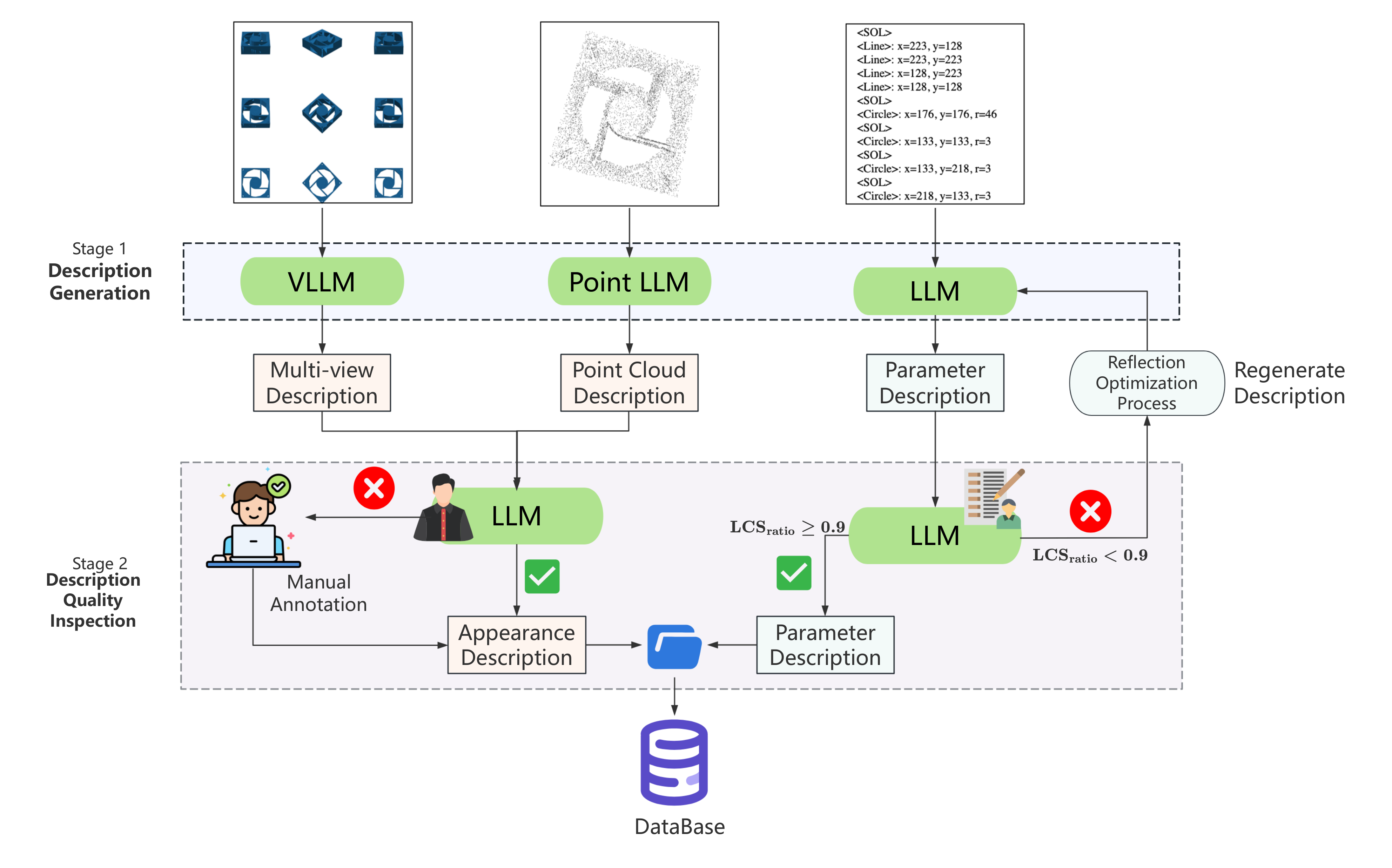}
    \caption{LLM-Based Semi-Automated Annotation for CAD Datasets Process. This pipeline primarily generates appearance and parameter descriptions for CAD datasets, focusing on the Description Generation and Description Quality Inspection stages.}
    \label{fig:LLM-Based Semi-Automated Annotation for CAD Datasets Process}
\end{figure*}

To efficiently create high-quality annotations for large CAD datasets, we propose a semi-automated annotation pipeline based on LLMs, utilizing multiple LLMs in a coordinated workflow. Annotating the appearance of CAD models requires human-machine collaboration, while parameter description annotation can be fully automated. Each annotation pipeline includes a description generation stage and a description quality control stage. The process is described in Fig.~\ref{fig:LLM-Based Semi-Automated Annotation for CAD Datasets Process}.

\textbf{(i). Semi-Automated Appearance Description Generation}

To address the challenge of efficiently labeling large-scale CAD datasets, we leverage LLMs to automate parameter descriptions and assist in appearance annotation. For the description generation stage, we first sample multiple views of the CAD model and use a VLLM to generate an appearance description based on these views. Simultaneously, point cloud data from the CAD model is processed by a PointLLM~\cite{xu2024pointllm} to generate descriptions based on the point cloud. The models answer key questions regarding the 3D model’s appearance features, material composition, specific details, and function.

For the description quality inspection stage,  we use LLM to check the consistency between the point cloud and multi-view descriptions and point out conflicts (e.g., one calling it a cylinder, another a cube). If the point cloud and multi-view descriptions align well, they are merged into a comprehensive appearance description; otherwise, manual annotation is required (our experiments showed an automatic pass rate of $98.4\%$, with only a few samples needing human intervention). The final appearance description comes from combining both the automated and manual annotations.

\textbf{(ii). Automated Parameter Description Generation}

The parameter description is primarily generated based on the CCS. 
For the description generation stage, annotated CCS parameter data is input into the LLM along with a predefined prompt template. The prompt also specifies that descriptions must be "fluent and instructive paragraphs" detailing each step of the CAD modeling process and the corresponding parameters.

For the description quality inspection stage, we validate the accuracy of the generated CCS parameter description by reverse verification. It is deemed correct if the model can reconstruct the ground truth CCS from the generated description. The reliability of the description is measured by the longest common subsequence (LCS) ratio between the generated CCS and the ground truth  CCS. The LCS metric effectively captures structural similarities between sequences while allowing for minor variations, making it well-suited for evaluating CAD modeling descriptions. To quantify the similarity between the generated description and the ground truth, we use the $\text{LCS}_{\text{ratio}}$:
\begin{align}
\text{LCS}_{\text{ratio}} = \frac{\text{len}(\text{LCS}(g,r))}{\text{len}(g)},
\label{eq:lcs_ratio}
\end{align}
where $g$ denotes the ground truth CCS and $r$ represents the generated CCS.
Descriptions with an $\text{LCS}_{\text{ratio}}$ above $0.9$ are accepted, while those with an LCS ratio below 0.9 are considered low reliability and enter a reflection optimization process.

\textbf{Reflection Optimization Process}: For low-reliability descriptions, we follow a reflection optimization process:
\begin{enumerate}
    \item Analyze potential issues in the description generation stage based on ground truth CCS.
    \item Generate a new CCS parameter description based on the model's reflection feedback from Step 1.
    \item Recheck the newly generated description.
\end{enumerate}

This process repeats until the description reaches the reliability threshold (i.e., $\text{LCS}_{\text{ratio}}$ $\ge$ $0.9$) or the maximum retry limit (twice). Note that this process is fully automated, requiring no human intervention for parameter description annotation.

\subsubsection{TCADGen}
\begin{figure*}[t]
\centering
  \includegraphics[width=0.9\textwidth]{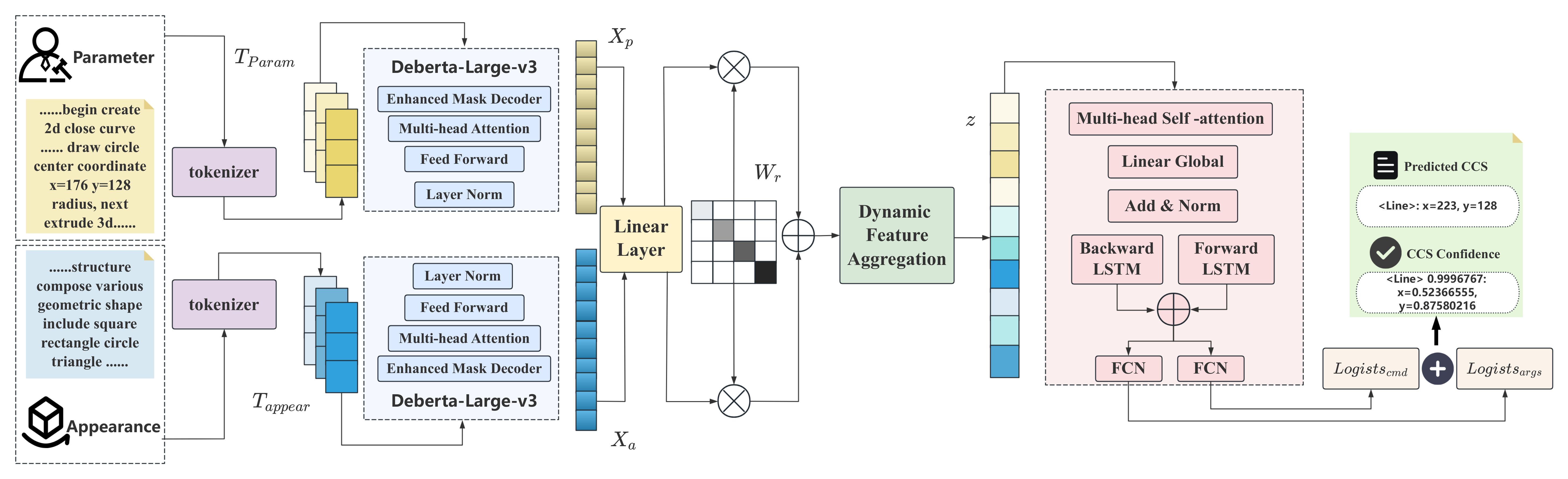}
  \caption{Transformer-based CAD Generator (TCADGen). }
  \label{fig:TCADGen}
\end{figure*}

We propose a novel CAD model generation framework called TCADGen, which innovatively adopts a dual-channel Transformer architecture to achieve an effective fusion of parameter modeling knowledge and visual appearance features, as shown in Fig.~\ref{fig:TCADGen}. 


\textbf{(i). Dual-Channel Feature Aggregator}
To fully preserve both parametric features and appearance features of the model, we propose a dual-channel feature aggregator that simultaneously embeds parameter descriptions and appearance descriptions. Parameter descriptions contain both basic geometric operations and specific parameter designs, while appearance descriptions include model appearance analysis. Based on DeBerta-Large-v3~\cite{deberta-tran2024deberta}, we first encode and project the input text through BERT and linear transformation:
\begin{align}
\mathbf{h}_p &= BERT(T_{param})\mathbf{W}_p + \mathbf{b}_p, \\
\mathbf{h}a &= BERT(T_{appear})\mathbf{W}_a + \mathbf{b}_a,
\end{align}
where $T_{param}$ and $T_{appear}$ are the input parameter and appearance description texts respectively, $d_p$ and $d_a$ are the dimensions of parameter and appearance features, $d$ is the dimension of the shared semantic space, $\mathbf{W}_p \in \mathbb{R}^{d_p \times d}$ and $\mathbf{W}_a \in \mathbb{R}^{d_a \times d}$ are projection matrices that map BERT embeddings into a shared $d$-dimensional semantic space, and $\mathbf{b}_p$, $\mathbf{b}_a$ represent bias vectors.

Inspired  by  capsule  networks~\cite{capsul-epatrick2022capsule}, we design an adaptive feature fusion mechanism based on dynamic routing to fully utilize the  normalized feature representations:
\begin{equation}
\mathbf{s}_j = \sum_i \text{softmax}(\mathbf{W}_r[\hat{\mathbf{h}}_p^i;\hat{\mathbf{h}}_a^j])\hat{\mathbf{h}}p^i\mathbf{W}{ij},
\end{equation}
where $\hat{\mathbf{h}}_p$ and $\hat{\mathbf{h}}_a$ denote the normalized representations of $\mathbf{h}_p$ and $\mathbf{h}_a$ respectively, $\mathbf{W}_r$ is a learnable routing weight matrix, $i$ and $j$ correspond to the subscripts of the feature tokens, $[\cdot;\cdot]$ represents feature concatenation, and $\mathbf{W}_{ij}$ is the feature transformation matrix between tokens $i$ and $j$.

The final fused features are obtained through a weighted combination:
\begin{equation}
\mathbf{z} = \sum_j \alpha_j\mathbf{s}_j,
\end{equation}
where $\alpha_j$ represents adaptive weight coefficients calculated through the attention mechanism and $\mathbf{z}$ is the final fused feature vector.

\textbf{(ii). CAD Command Sequence Decoder}

We then decode the fused feature vector $\mathbf{z}$ into a standardized CAD modeling sequence. We first use a multi-head attention mechanism to capture global  dependencies:
\begin{equation}
\mathbf{H}^l = \text{MultiHead}(\mathbf{W}_Q^l\cdot\text{PE}(\cdot), \mathbf{W}_K^l\cdot\mathbf{z}, \mathbf{W}_V^l\cdot\mathbf{z}),
\end{equation}
where $\mathbf{H}^l$ is the output of the $l$-th attention layer, $\mathbf{W}_Q^l$, $\mathbf{W}_K^l$, and $\mathbf{W}_V^l$ are learnable query, key, and value projection matrices respectively, and $\text{PE}(\cdot)$ represents positional encoding. To better capture the sequential modeling patterns, we further incorporate  bidirectional LSTM~\cite{lstm-greff2016lstm} into our architecture:
\begin{equation}
\mathbf{out}_t = [LSTM(\overrightarrow{\mathbf{h}}_t); LSTM(\overleftarrow{\mathbf{h}}_t)],
\end{equation}
where $\mathbf{out}_t$ is the output at time step $t$, $\overrightarrow{\mathbf{h}}_t$ and $\overleftarrow{\mathbf{h}}_t$ denote the forward and backward hidden states respectively, and $[\cdot;\cdot]$ represents concatenation.

For efficient sequence generation,  we design  the model to predict the entire CAD modeling sequence in parallel:
\begin{equation}
p(\widehat{\mathbf{M}}|\mathbf{z}, \Theta) = \prod_{i=1}^{N_c} p(\mathbf{t}_i, \widehat{\mathbf{p}}_i|\mathbf{z}, \Theta),
\end{equation}
where $\widehat{\mathbf{M}}$ represents the predicted CAD modeling sequence, $\Theta$ denotes the model parameters, $N_c$ is the number of CAD commands in the sequence, $\mathbf{t}_i$ represents the command type at position $i$, and $\widehat{\mathbf{p}}_i$ represents the predicted parameters for the command at position $i$.

\subsubsection{LLM-Enhanced CAD Command Sequence Generation}

The results predicted by TCADGen can assist LLM in generating more accurate CCS. In this paper, we propose an enhanced CCS generation process based on large language models, referred to as CADLLM. This process combines the CCS generated by TCADGen, prediction confidence information, and user-provided description, enabling the precise generation of CCS through a fine-tuned large language model. 

During the training phase, the fine-tuned model is trained on 1,000 samples (the choice of dataset size is explained in the experimental section, RQ3) that are consistent with the distribution of the dataset. The task is designed such that the output of TCADGen (predicted CCS and its confidence) serves as the query for supervised fine-tuning (SFT) to generate the correct CCS as the response. By learning the mapping between CCS and confidence, CADLLM corrects the errors generated by TCADGen and produces more accurate CCS.

During the inference phase, CADLLM acts as the inference model. It takes as input the user’s parameters, appearance description, CCS generated by TCADGen based on user requirements, and the confidence information and outputs the enhanced CCS data.
\section{Experiments}


In our experiments, we address three research questions to evaluate the effectiveness of the proposed methodologies. For a detailed description of the dataset and parameters, please refer to Appendix~\ref{sec:Dataset} and ~\ref{sec:Parameter Setting}.

\noindent\hspace{1em}\textbf{RQ1: How effective is the proposed LLM-based semi-automated annotation framework in improving the quality of annotations for CAD datasets?}\label{RQ1}

We have proposed a human-machine semi-automated CAD dataset annotation process that utilizes large models for annotation and verification to ensure quality. For the parameter description generation task, both the description generation and describe quality inspection stages utilize the gemma-2-27b-it~\cite{google_gemma_2_27b_it} model (the reasons for model selection and parameter description generation evaluation are provided in Appendix~\ref{sec:LLMs to Generate Parameter Descriptions}). The reflection optimization process uses the gemma-2-27b-it model. For the appearance description generation task, the VLLM used for the description generation stage is the Llama-3.2-11B-Vision-Instruct~\cite{meta_llama_3_2_11b_vision_instruct} model, while the point cloud model uses PointLLM~\cite{xu2024pointllm}. The description quality inspection stage also utilizes the gemma-2-27b-it model. 
\begin{table*}[ht]
\resizebox{\textwidth}{!}{
\begin{tabular}{lccc|cccccccc}
\hline
\multirow{3}{*}{\textbf{Models}} &
  \multicolumn{3}{c|}{\multirow{2}{*}{\textbf{Avg Command}}} &
  \multicolumn{8}{c}{\textbf{Each Command}} \\
\cline{5-12}
 &
   &
   &
   &
  \multicolumn{2}{c}{Line} &
  \multicolumn{2}{c}{Arc} &
  \multicolumn{2}{c}{Circle} &
  \multicolumn{2}{c}{Extrude} \\
\cline{2-12}
 &
  ACC &
  F1 &
  AUC &
  AUC &
  F1 &
  AUC &
  F1 &
  AUC &
  F1 &
  AUC &
  F1 \\ 
\hline
DeepCAD &
  0.571 &
  0.606 &
  0.747 &
  0.648 &
  0.797 &
  0.540 &
  0.588 &
  0.587 &
  0.551 &
  0.616 &
  0.615 \\
Text2CAD &
  0.840 &
  0.722 &
  0.819 &
  0.763 &
  0.904 &
  0.584 &
  0.601 &
  0.751 &
  0.701 &
  0.772 &
  0.705 \\
BERT fine-tuned w/o &
  0.807 &
  0.731 &
  0.828 &
  0.760 &
  0.937 &
  0.637 &
  0.690 &
  0.757 &
  0.690 &
  0.711 &
  0.622 \\
Dual-channel w/o &
  0.847 &
  0.769 &
  0.850 &
  0.791 &
  0.945 &
  0.668 &
  0.523 &
  0.802 &
  0.762 &
  0.746 &
  0.684 \\
TCADGen(Text2CAD-dataset)&
  0.804 &
  0.731 &
  0.822 &
  0.760 &
  0.920 &
  0.613 &
  0.590 &
  0.780 &
  0.731 &
  0.795 &
  0.767 \\
\hline
\textbf{TCADGen} &
  0.890 &
  0.771 &
  0.854 &
  0.808 &
  0.950 &
  0.682 &
  0.642 &
  0.837 &
  0.771 &
  0.781 &
  0.746 \\ 
\textbf{TCADGen+CADLLM (Ours)} &
  \textbf{0.966} &
  \textbf{0.947} &
  \textbf{0.962} &
  \textbf{0.957} &
  \textbf{0.979} &
  \textbf{0.925} &
  \textbf{0.924} &
  \textbf{0.959} &
  \textbf{0.960} &
  \textbf{0.942} &
  \textbf{0.946} \\ 
\hline
\end{tabular}
}
\caption{Comparison of different models and datasets for CAD command prediction. The Average Command section shows the overall performance metrics, while the Per Command section shows detailed AUC and F1 scores for a specific command type. Bold numbers indicate the best performance.}
\label{tab:RQ2}
\end{table*}

The experimental results demonstrate that this framework significantly improves annotation quality over the Text2CAD baseline~\cite{khan2025text2cad} across most evaluation metrics.The experimental results are presented in Table~\ref{tab:RQ2}, with the TCADGen (Text2CAD-dataset) row displaying the corresponding results. For overall performance, command line accuracy improves from $0.804$ to $0.890$, an increase of about $8.6$ percentage points. The framework shows enhanced performance on commands such as Arc, Circle, and Extrude.  Notably, the  AUC-F1 score for Circle operations increases from  $0.731$ to $0.771$, reflecting improved annotation capability for circular geometric elements. However, performance on complex operations, such as Extrude,  shows a slight decrease from $0.767$ to $0.746$, suggesting areas for further improvement.

\begin{figure}[h]
\includegraphics[width=0.8\columnwidth]{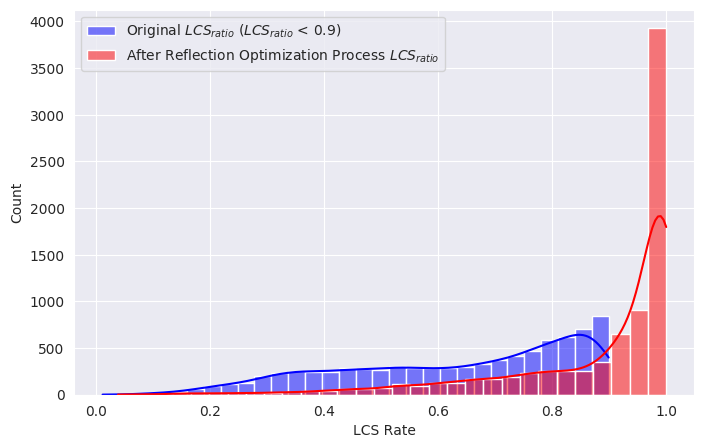}
    \caption{Distribution of $\text{LCS}_{\text{ratio}}$ scores before and after the Reflection Optimization Process. }
  \label{fig:Reflection Optimization score}
\end{figure}

\begin{figure}[h]
\includegraphics[width=0.8\columnwidth]{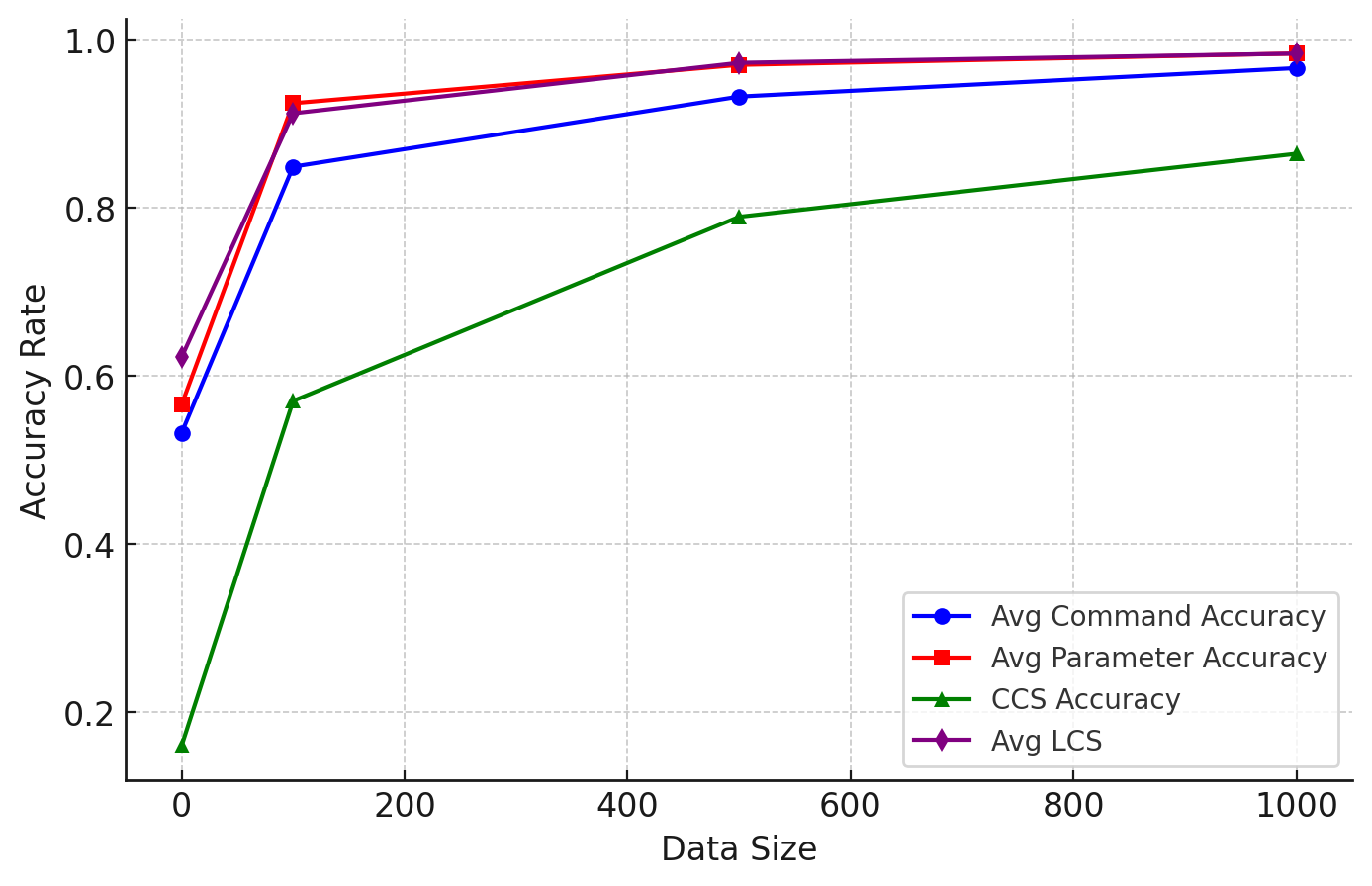}
  \caption{The impact of data size on CADLLM performance. As the data size increases from 0 to 1000 samples.
}
  \label{fig:data—size}
\end{figure}

Additionally, we assess the reflection optimization process for enhancing CCS parameter descriptions with low quality ($\text{LCS}_{\text{ratio}}$ < 0.9). This process, which includes error analysis, reflective feedback, and re-evaluation, shows a significant increase in confidence scores ($t = -137.76, p < 0.001$), confirming its effectiveness. Histogram and Kernel Density Estimate (KDE) analyses (see Fig.~\ref{fig:Reflection Optimization score}) reveal that after optimization, $\text{LCS}_{\text{ratio}}$ scores cluster near $1.0$, indicating improved reliability and consistency.



\begin{table*}[h]
\centering
\small
\setlength{\tabcolsep}{8pt}  
\begin{tabular}{l ccc}
\hline
\textbf{Input} & \textbf{Model} & \textbf{Acc.} & \textbf{Avg $\text{LCS}_{\text{ratio}}$} \\
\hline
\multirow{3}{*}{\begin{tabular}[t]{@{}l@{}}Parameter Description and\\Appearance Description\end{tabular}} 
& GPT-4o(prompt) & 0.617 & 0.977 \\
& Llama 3.2 3B(prompt) & 0.328 & 0.688 \\
& Llama 3.2 3B(SFT) & 0.621 & 0.883 \\
\hline
\multirow{6}{*}{\begin{tabular}[t]{@{}l@{}}Transformer-based Prediction\\of CCS and CCS Confidence\end{tabular}} 
& GPT-4o(prompt) + TCADGen & 0.670 & 0.947 \\
& Claude-3.5(prompt) + TCADGen & 0.812 & 0.963 \\
& Llama 3.2 3B(prompt) + TCADGen & 0.328 & 0.683 \\
& CADLLM + DeepCAD & 0.677 & 0.978 \\
& CADLLM + Text2CAD & 0.710 & 0.974 \\
& \textbf{CADLLM + TCADGen(ours)} & \textbf{0.864} & \textbf{0.983} \\
\hline
\end{tabular}
\caption{Performance comparison of different methods for CAD command sequence generation. Our proposed CADLLM + TCADGen achieves the best performance in both accuracy and average $\text{LCS}_{\text{ratio}}$.}
\label{tab:RQ3_table1}
\end{table*}

\begin{table*}
\centering
\small
\setlength{\tabcolsep}{4.2pt}
\begin{tabular}{ccccc}
\hline
\textbf{Method} & \textbf{Model} & \textbf{CD}$\downarrow$ & \textbf{MMD}$\downarrow$ & \textbf{JSD}$\downarrow$ \\ 
\hline
\multirow{4}{*}{Transformer-based} & DeepCAD & 169.93 & 31.91 & 45.03 \\
& Text2CAD & 142.83 & 28.98 & 40.23 \\
& CAD Translator & - & 2.94 & 10.92 \\
& TCADGen & 120.99 & 21.36 & 35.25 \\
\hline
\multirow{4}{*}{LLM-based} & CADFusion & 45.67 & 3.49 & 17.11 \\
& DeepCAD+CADLLM & 4.25 (\textcolor{red}{$\downarrow$ 165.68}) & 3.13 (\textcolor{red}{$\downarrow$ 28.78}) & 8.58 (\textcolor{red}{$\downarrow$ 36.45}) \\
& Text2CAD+CADLLM & 4.31 (\textcolor{red}{$\downarrow$ 138.52})& 3.12 (\textcolor{red}{$\downarrow$ 25.86}) & 8.42 (\textcolor{red}{$\downarrow$ 31.81}) \\
& TCADGen+CADLLM(ours) & \textbf{3.12} (\textcolor{red}{$\downarrow$ 117.87})& \textbf{2.78}(\textcolor{red}{$\downarrow$ 18.58}) & \textbf{8.38} (\textcolor{red}{$\downarrow$ 26.87}) \\ 
\hline
\end{tabular}
\caption{Quantitative comparison of Chamfer Distance (CD), Minimum Matching Distance (MMD), and Jensen-Shannon Divergence (JSD) metrics, where lower values indicate better performance (↓).}
\label{tab:CD}
\end{table*}

\noindent\hspace{1em}\textbf{RQ2: How well does TCADGen perform in translating the human modeling language into standard CAD command sequences?}\label{RQ2}

As shown in Table ~\ref{tab:RQ2}, the experimental results demonstrate that TCADGen significantly outperforms existing methods in generating CAD modeling sequences. Compared to the DeepCAD baseline, TCADGen achieves a $31.8$ percentage point improvement in average command accuracy and surpasses DeepCAD in overall performance metrics, with an $F1$ score of $0.771$ and AUC of $0.854$. Additionally, in comparison to Text2CAD, TCADGen shows a $5$ percentage point improvement in average command accuracy and demonstrates better performance for complex modeling commands, such as Line (AUC from $0.648$ to $0.808$), Arc (AUC from $0.540$ to $0.682$), and Circle (AUC from $0.587$ to $0.837$). The ablation studies further validate this effectiveness, showing that removing CAD-specific BERT fine-tuning leads to a $4$ percentage points drop in performance, while the dual-channel architecture significantly improves complex command generation, e.g., Extrude ($F1 = 0.7457$), indicating better distribution alignment and semantic consistency with ground truth commands.




\noindent\hspace{1em}\textbf{RQ3: How effective are LLM in correcting TCADGen-generated CAD command sequences?}\label{RQ3}

We compare two methods of using LLMs to generate CCS: one that directly generates CCS using parameter and appearance descriptions and the other that combines transformer-based predictions with CCS confidence scores. Even with fine-tuning, the direct description method, such as Llama 3.2 3B with prompt or LoRA, performs worse than the Transformer-based approach. For example, Llama achieves $32.8\%$ accuracy ($\text{LCS}_{\text{ratio}}$ $0.6881$), while GPT-4o with predictions and confidence improves to $67.0\%$ accuracy ($\text{LCS}_{\text{ratio}}$ $0.9477$). The optimal method, CADLLM + TCADGen, achieves $86.4\%$ accuracy and $\text{LCS}_{\text{ratio}}$ of $0.983$, demonstrating that combining Transformer predictions with confidence scores is significantly more effective than direct generation from descriptions.

The evaluation results are presented in Table~\ref{tab:CD} for model generation quality. Among Transformer-based methods, TCADGen achieves the best performance, with a Chamfer Distance (CD) of 120.99, Minimum Matching Distance (MMD) of 21.36, and Jensen-Shannon Divergence (JSD) of 35.25. In comparison, CAD Translator~\cite{CADranslator}, which employs a one-stage framework, shows inferior performance, with a CD of 142.83. The introduction of CADLLM significantly enhances the performance of Transformer-based models. Specifically, TCADGen+CADLLM (ours) achieves a CD of 4.52 (↓116.47), MMD of 2.78 (↓18.17), and JSD of 8.38 (↓26.57). Meanwhile, CADFusion~\cite{textTocad}, which utilizes a two-stage training process with LLaMA-3-8b-Instruct, demonstrates lower accuracy, with a CD of 45.67. In contrast, our approach, which employs the smaller LLaMA-3.2-3b-Instruct model, outperforms both CADFusion and CAD Translator in all key metrics, further highlighting the effectiveness of CADLLM.

Our experimental results show that increasing the fine-tuning data size (from 0 to 1000 samples) leads to substantial performance improvements in CADLLM (Fig.~\ref{fig:data—size}, Table~\ref{tab:RQ3-data}). Specifically, CCS accuracy improves from $16.0\%$ to $86.4\%$, while $\text{LCS}_{\text{ratio}}$ increases from $0.622$ to $0.983$. The performance improves only slightly after 500 samples, showing that 1000 samples achieve the best model performance and training cost balance.

\section{Conclusion}
This paper proposes a CAutoD framework that combines LLMs with domain-specific CAD parametrization to generate CAD models from user descriptions in the detailed design phase. We have demonstrated that, owing to the key innovations such as a semi-automated data annotation pipeline, TCAD-Gen, and LLM-assisted enhancement, the framework has significantly outperformed the existing methods in CAD modeling sequence generation accuracy and efficiency, making it a valuable tool for generating precise CAD models from textual prompts. Future work will focus on scaling the framework and exploring its applications in broader industrial design scenarios.

\section*{Limitations}
Our framework demonstrates promising results in generating CAD modeling sequences, but several challenges remain. The semi-automatic data annotation process is resource-intensive, requiring a large number of LLM calls and manual verification for quality control. While automation reduces human effort, inconsistencies between multi-view descriptions and point cloud representations still require intervention, limiting scalability for large datasets.

The imbalance in command distributions affects model robustness, as certain operations, such as "Line," appear significantly more often than others like "Arc." This bias in training data leads to better performance on frequent commands while limiting generalization to complex 3D geometries that require underrepresented operations.

While the framework effectively generates and optimizes CAD sequences, it does not explicitly incorporate geometric constraints or structural reasoning, resulting in syntactically correct sequences that may not always align with practical design requirements. Future work could explore integrating geometric priors and constraint-aware learning to improve the reliability and applicability of automated CAD modeling.

Our framework focuses on the detailed design process during the CAD design phase and does not provide adequate support for the conceptual design phase, where parameter descriptions may be incomplete or vague.




\section*{Acknowledgements}
This work is supported by the National Natural Science Foundation of China (Grant No. 52405253), by the Ministry of Science and Technology of China (Grant No. H20240917), and by the Sichuan Province Science and Technology Innovation Seedling Project (Grant No. MZGC20240134).

\clearpage
\bibliographystyle{acl}
\bibliography{reference}

\clearpage  
\appendix
\section{Appendix}
\label{sec:appendix}
\subsection{CAD Command Sequence}\label{sec:CCS_introduction}
\begin{figure*}[t]    
    \centering
    \includegraphics[width=1\textwidth]{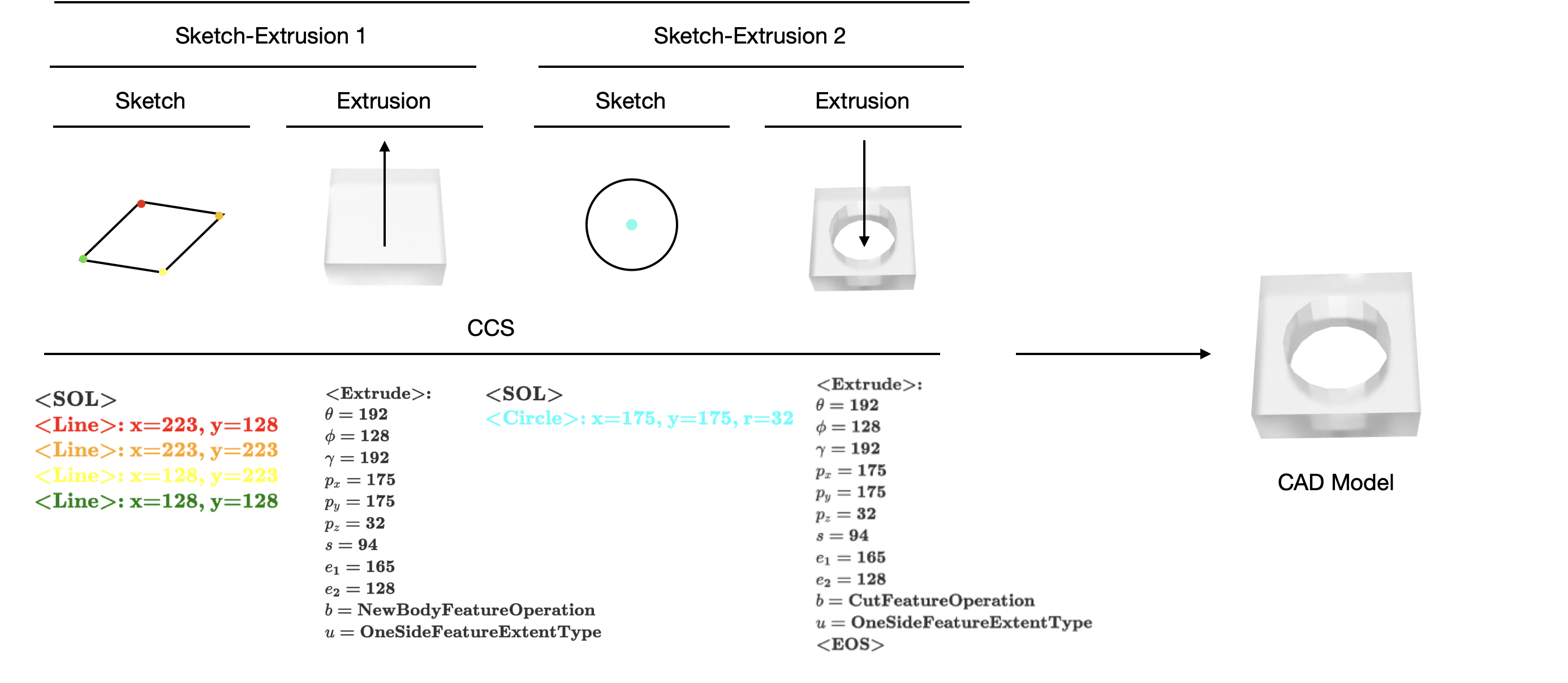}
    \caption{Illustration of the CAD Command Sequence (CCS) representation. The process shows two sketch-extrusion operations: (1) creating a base by sketching a rectangle using Line commands with specified coordinates, followed by an extrusion operation to form a 3D block; (2) adding a circular hole by sketching a circle and applying a cut extrusion operation. }
    \label{fig:ccs-workflow}
\end{figure*}

In the CAD modeling task proposed in this paper, we adopt a command sequence representation based on sketch-extrusion, and the generated sequence is referred to as the CAD Command Sequence (CCS). The CCS example is shown in Fig.~\ref{fig:ccs-workflow}, and the detailed CCS parameters are presented in Table~\ref{tab:ccs_parameters}.

The sequence consists of two-dimensional sketch commands and three-dimensional extrusion commands. Specifically, the two-dimensional sketch section begins with the tag \texttt{<SOL>}, representing the start of a closed contour. Each contour can be constructed using three basic curve commands:

\begin{itemize}
    \item The \textbf{Line} command determines the direction and length of the straight line by specifying the endpoint coordinates \((x, y)\);
    \item The \textbf{Arc} command requires the endpoint coordinates \((x, y)\), sweep angle \(\alpha\), and direction flag \(f\) to define the shape and direction of the arc;
    \item The \textbf{Circle} command creates a complete circular contour by specifying the center coordinates \((x, y)\) and radius \(r\).
\end{itemize}

Once the two-dimensional sketch is completed, the \textbf{Extrude} command \(E\) transforms it into a three-dimensional solid. This command includes several key parameters:

\begin{itemize}
    \item The spatial orientation of the sketch plane is determined by the Euler angles \((\theta, \phi, \gamma)\);
    \item The translation vector \((p_x, p_y, p_z)\) specifies the position of the origin of the sketch plane;
    \item The scaling factor \(s\) controls the size of the contour;
    \item The bidirectional extrusion distances \((e_1, e_2)\) determine the thickness of the solid;
    \item The Boolean operation type \(b\) is used to specify the interaction with existing geometry, which can be a new entity, union, subtraction, or intersection operation;
    \item The extrusion type \(u\) specifies the extrusion method, including unidirectional, symmetric, or bidirectional extrusion.
\end{itemize}

The CAD sequence can be modeled by creating two-dimensional sketches and repeatedly applying extrusion commands.

\begin{table*}[t]

\renewcommand{\arraystretch}{1.3}
\centering
\begin{tabular}{lc@{\hspace{2em}}cp{7cm}}
\toprule
\textbf{Command} & \textbf{Parameter} & \textbf{Value Range} & \textbf{Description} \\ 
\midrule
\texttt{<SOL>} & \(\emptyset\) & - & Contour Start Marker \\ 
\midrule
\textbf{Line} & \(x\) & \([0,255]\) & End point x coordinate of the line segment \\ 
              & \(y\) & \([0,255]\) & End point y coordinate of the line segment \\ 
\midrule
\textbf{Arc}  & \(x\) & \([0,255]\) & End point x coordinate of the arc \\ 
              & \(y\) & \([0,255]\) & End point y coordinate of the arc \\ 
              & \(\alpha\) & \([0,255]\) & Sweep angle of the arc \\ 
              & \(f\) & \(\{0,1\}\) & Counterclockwise direction flag \\ 
\midrule
\textbf{Circle} & \(x\) & \([0,255]\) & Center x coordinate of the circle \\ 
                & \(y\) & \([0,255]\) & Center y coordinate of the circle \\ 
                & \(r\) & \([0,255]\) & Radius of the circle \\ 
\midrule
\textbf{Extrude} & \(\theta\) & \([0,255]\) & Rotation angle around the x-axis of the sketch plane \\ 
                 & \(\phi\) & \([0,255]\) & Rotation angle around the y-axis of the sketch plane \\ 
                 & \(\gamma\) & \([0,255]\) & Rotation angle around the z-axis of the sketch plane \\ 
                 & \(p_x\) & \([0,255]\) & x-coordinate of the sketch plane origin \\ 
                 & \(p_y\) & \([0,255]\) & y-coordinate of the sketch plane origin \\ 
                 & \(p_z\) & \([0,255]\) & z-coordinate of the sketch plane origin \\ 
                 & \(s\) & \([0,255]\) & Scaling factor for the sketch contour \\ 
                 & \(e_1\) & \([0,255]\) & Positive direction extrusion distance \\ 
                 & \(e_2\) & \([0,255]\) & Negative direction extrusion distance \\ 
                 & \(b\) & \(\{7,8,9,10\}\) & Boolean operation type: \\[0.3em]
                 & & & 7: NewBodyFeatureOperation \\[0.2em]
                 & & & 8: JoinFeatureOperation \\[0.2em]
                 & & & 9: CutFeatureOperation \\[0.2em]
                 & & & 10: IntersectFeatureOperation \\[0.3em] 
                 & \(u\) & \(\{1,2,3\}\) & Extrusion type: \\[0.3em]
                 & & & 1: OneSideFeatureExtentType \\[0.2em]
                 & & & 2: SymmetricFeatureExtentType \\[0.2em]
                 & & & 3: TwoSidesFeatureExtentType \\ 
\midrule
\texttt{<EOS>} & \(\emptyset\) & - & Sequence End Marker \\ 
\bottomrule
\end{tabular}
\caption{CAD Command Sequence (CCS) Parameters.}
\label{tab:ccs_parameters}

\end{table*}

\subsection{Dataset}\label{sec:Dataset}
The DeepCAD dataset~\cite{deepcad-wu2021deepcad}, derived from the ABC dataset~\cite{koch2019abc}, is a large-scale collection of 178,000 parametric CAD models designed for CAD model generation and reconstruction. Unlike other datasets, such as Fusion 360~\cite{fusion360-willis2021fusion} with only 8,000 instances, DeepCAD includes sketch-extrusion sequences, enabling models to learn procedural modeling behaviors rather than static geometry. Its extensive cross-industry coverage ensures strong generalization to external datasets, reducing concerns about overfitting. For this study, DeepCAD was preprocessed to ensure consistency, remove incomplete instances, and improve training efficiency. The corresponding STL file was generated for each sample based on the original model information provided in the dataset's JSON files. Simultaneously, the CCS descriptions were extracted from the JSON files. The generated STL files were then sampled from nine different viewpoints to produce multi-view images. Additionally, point sampling was performed files, generating point cloud files containing 8,000 points. The subsequent training and validation phases included only those samples for which the STL files, multi-view images, and point cloud files were successfully generated. The final dataset used in this experiment consisted of 155,503 training samples and 5,647 test samples.

\subsection{Parameter Setting}\label{sec:Parameter Setting}

We use two main parts in our system: a TCADGen model and a fine-tuned LLM(CADLLM).

For TCADGen, we build a simple Transformer decoder. It has $4$ layers and $8$ attention heads in each layer. The model's base size is $d_{model}=256$, and it can handle sequences up to $300$ tokens. The argument size is $256$, and the feedforward layers are $512$ units wide. We use a dropout of $0.1$ throughout the model to avoid overfitting. We trained the model using a single A6000 GPU for 24 hours. The inference of 5,647 data samples takes 240 seconds.

For the LLM part, Our system uses two main parts and fine-tune it using LoRA, leveraging the LlamaFactory framework~\cite{zheng2024llamafactory} for fine-tuning. We set the basic batch size to $1$ and use gradient accumulation for $8$ steps. This gives us an effective batch size of $8$. We use a learning rate of $1e^{-4}$ with cosine decay and $10\%$ warmup steps. We train the model for $3$ epochs with BF16 precision. This helps us save memory and train faster. The model can take inputs up to $4096$ tokens long. We check the model's performance every $500$ step using $20\%$ of our data as a validation set. We save the model version that performs best. We trained and performed inference using a single A6000 GPU. The training duration was 2 hours. For inference, we utilized the vllm framework~\cite{kwon2023efficient} with a concurrency of 2. Inference of 5,647 test samples took 1.5 hours

\begin{figure*}[t]
    \centering
    \includegraphics[width=1\linewidth]{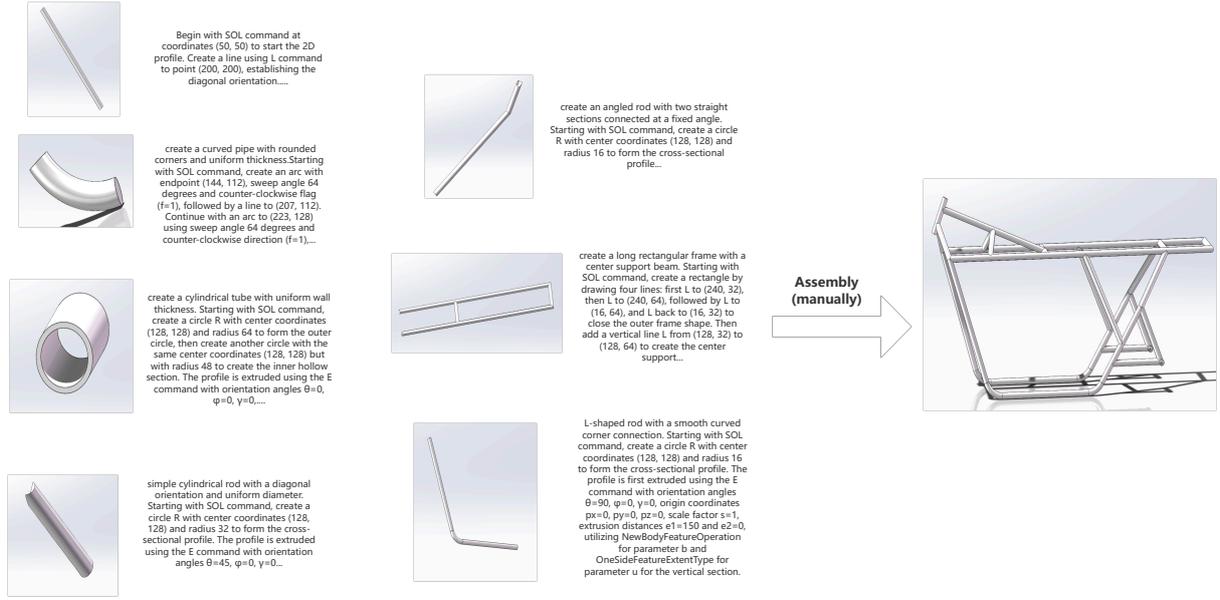}
    \caption{The motorcycle frame design process: Parts generated using our framework are manually assembled to create the final frame. The figure shows the individual components and their integration into the complete design.}
    \label{fig:Motorcycle Frame}
\end{figure*}
\subsection{Application of the Framework for Motorcycle Frame Model Generation}

In this work, we propose a framework aimed at detailed CAD design, where both the parameters and the full specifications of the model are predefined. The framework generates individual components of the motorcycle frame, which are then manually assembled to produce the final CAD model. Specifically, the proposed framework facilitates the rapid generation of parts that conform to the specified requirements, based on textual descriptions of both appearance and parameters. The generated models are presented in Fig.~\ref{fig:Motorcycle Frame}.

\subsection{Examples of annotation processes}
\subsubsection{Parameter Description Labeling Process}

\begin{figure*}
    \centering
    \includegraphics[width=1\linewidth]{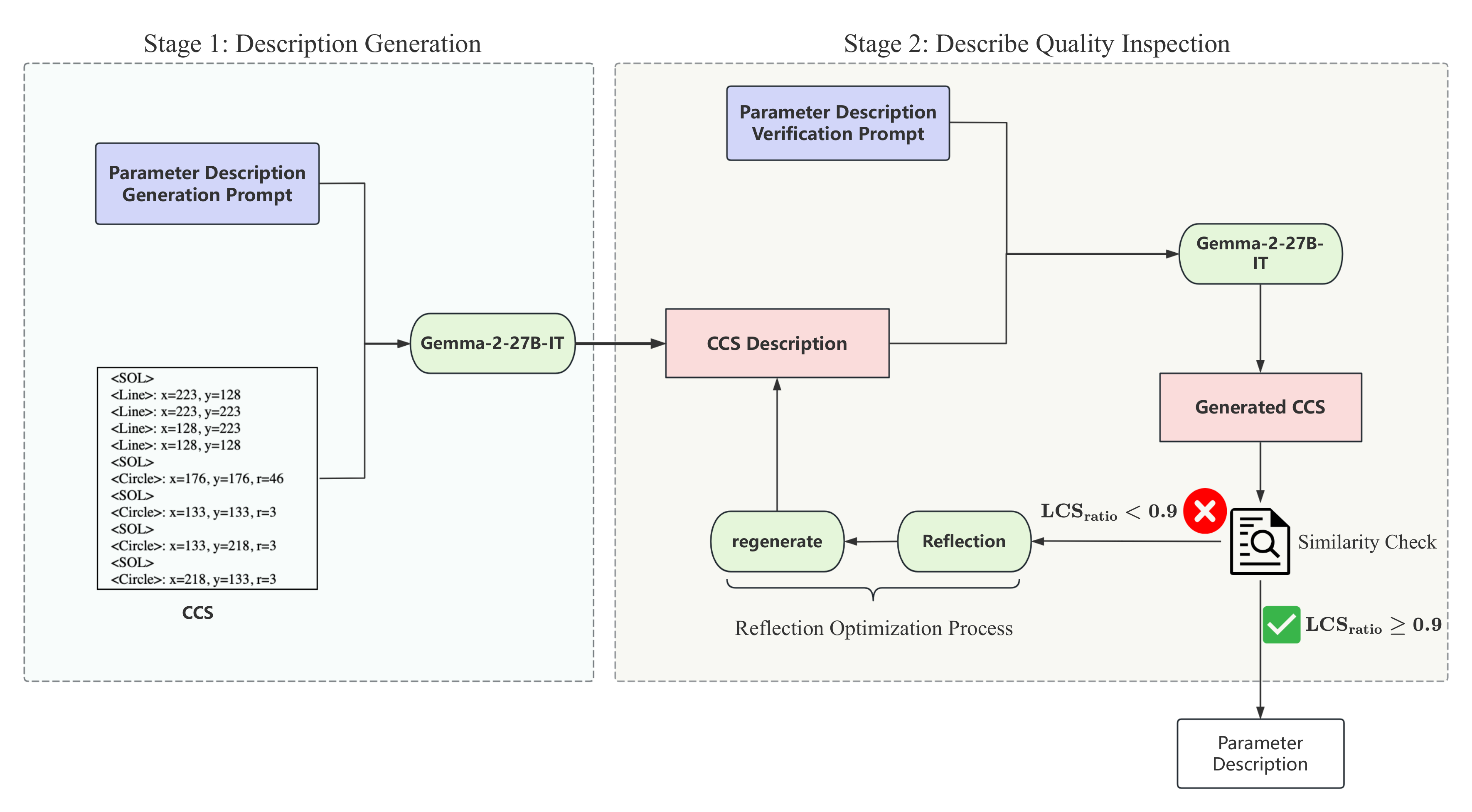}
\caption{Parameter Description Annotation Process Flow.} \label{fig:Parameter_description_labeling}
\end{figure*}

Figure~\ref{fig:Parameter_description_labeling} shows the Parameter Description Labeling Process.Table~\ref{tab:param_desc_examples} shows the parameter descriptions generated by the annotation process proposed in this paper, where $\text{LCS}_{\text{ratio}}$ refers to the comparison between the CCS generated by the large language model in the description quality control stage and the ground truth. When the $\text{LCS}_{\text{ratio}}$ of a description (e.g., No.00159955) is greater than 0.9, it indicates that the parameter description is successfully synthesized and can be adopted. If the $\text{LCS}_{\text{ratio}}$ is below 0.9 (e.g., No.00262583), the description enters the subsequent reflection and optimization process.

\begin{table*}[t]
\renewcommand{\arraystretch}{1.3}

\centering
\small
\resizebox{\textwidth}{!}{
\begin{tabular}{@{}p{2cm}p{8cm}cp{4cm}@{}}
\toprule
\textbf{ID} & \makecell[c]{\textbf{Generated}\\\textbf{Parameter Description}} & \textbf{$\text{LCS}_{\text{ratio}}$} & \textbf{Model Visualization} \\
\midrule
00159955 & \parbox{8cm}{\centering The sequence begins with creating two concentric circles to define the initial 2D closed curve. The first circle is centered at coordinates (176, 128) with a radius of 47, and the second circle, also centered at the same point, has a smaller radius of 40. This closed curve is then extruded with the extrusion parameters $\theta$=128, $\varphi$=128, and $\gamma$=128, indicating the orientation of the sketch plane. The sketch plane's origin is set at (70, 128, 128), and a scale factor of 115 is applied to adjust the profile size. The extrusion distances are set to e1=142 and e2=128, specifying how far the shape extends in the extrusion direction. The operation type for the extrusion is "NewBodyFeatureOperation," meaning a new body is created as a result. The extrusion direction is one-sided, indicated by the parameter u=OneSideFeatureExtentType, ensuring the shape is extruded in one direction.

Next, a new 2D closed curve is created with a combination of lines and arcs. Starting from the point (140, 128), a line is drawn to (195, 128), followed by an arc that sweeps counter-clockwise from (195, 128) to (195, 223) with a sweep angle $\alpha$=128. A series of lines are then drawn to complete the curve, and two circles are added at the coordinates (128, 140) and (128, 211), each with a radius of 6. These circles are also extruded using the parameters $\theta$=128, $\varphi$=128, and $\gamma$=128, with the origin set at (46, 70, 128) and a scale factor of 115. The extrusion distances are e1=142 and e2=128, with a "JoinFeatureOperation" applied to merge this new feature with the existing body. The extrusion is performed in a one-sided direction, ensuring the shape extends only in one direction...

} & 0.99 & \begin{center}\includegraphics[width=3.5cm]{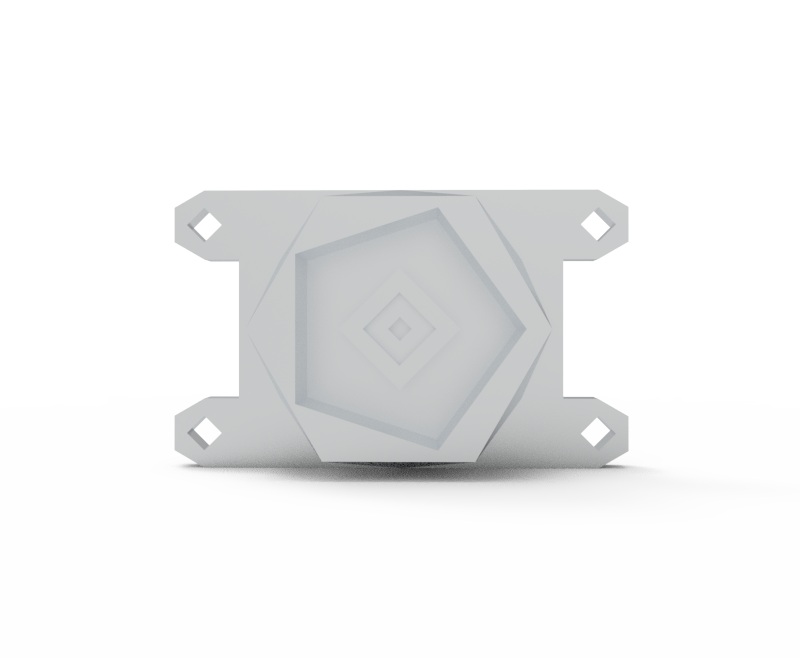}\end{center} \\
\midrule
00262583 & \parbox{8cm}{\centering The modeling process begins with the creation of two circles, both centered at (176, 128) with a radius of 48. The first circle is extruded into a 3D feature with an extrusion direction defined by the parameters $\theta$=128, $\varphi$=128, and $\gamma$=128, and the origin placed at (56, 170, 128). The extrusion uses a scale factor of 12 and extends by 176 units in the first phase and 128 units in the second, creating a new body i, extendingion. The second circle is then extruded using similar parameters, but the origin shifts to (188, 86, 128), and a "JoinFeatureOperation" is applied to merge the extrusion with the existing feature.

Next, a more complex 2D closed curve is created using arcs and lines. The curve is defined with specific arc directions and sweep angles, forming a polygon that closes at the starting point. A circle with a radius of 4 is defined at (132, 128) and extruded using the same extrusion parameters, extending the shape into 3D...
} & 0.74 & \begin{center}\includegraphics[width=3.5cm]{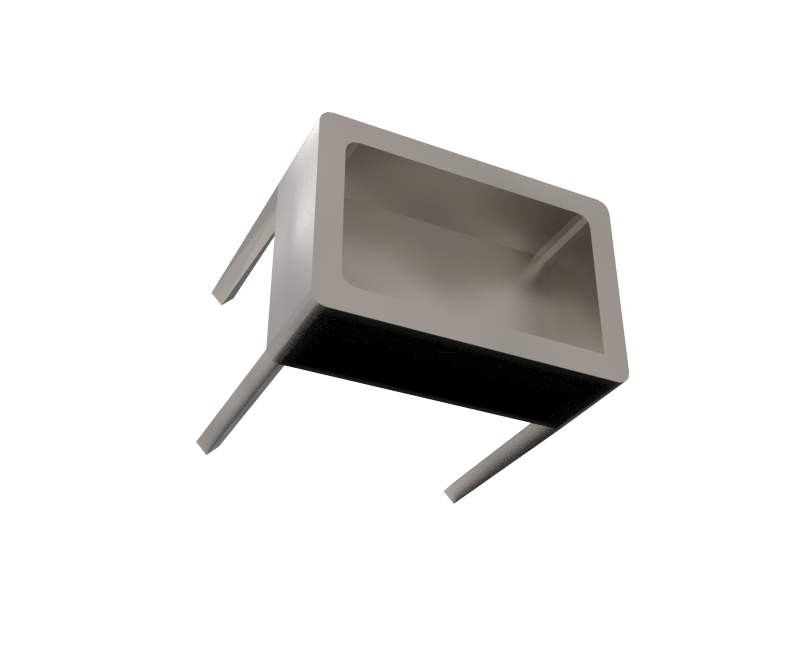}\end{center} \\
\bottomrule
\end{tabular}
}
\caption{Examples of parameter description generation. }
\label{tab:param_desc_examples}

\end{table*}

\paragraph{Reflection Optimization Process}

The original CCS and the optimized descriptions are shown in Table~\ref{tab:comparison}. The reflection optimization process is detailed in Table~\ref{tab:reflection-results}, where, shows the original CCS and optimized descriptions after two rounds of optimization, the description of the generated parameters achieves an $\text{LCS}_{\text{ratio}}$ of 1.

\begin{table*}[t]

\tiny
\centering
\begin{tabular}{p{5cm}p{7cm}c}
\toprule
\textbf{Original CCS} & \textbf{Generated Parameter Description} & \textbf{$\text{LCS}_{\text{ratio}}$} \\
\midrule
\texttt{<SOL>} & First, draw an arc with an end-point at x=144, y=112, & \multirow{19}{*}{0.675} \\
<Arc>: x=144, y=112, $\alpha$=64, f=1 & a sweep angle of 64 degrees, and a counter-clockwise & \\
<Line>: x=207, y=112 & direction. Then, draw a line to x=207, y=112, & \\
<Arc>: x=223, y=128, $\alpha$=64, f=1 & followed by an arc to x=223, y=128 with a sweep & \\
<Line>: x=223, y=204 & angle of 64 degrees and a counter-clockwise direction. & \\
<Arc>: x=207, y=220, $\alpha$=64, f=1 & Next, draw a line to x=223, y=204, an arc to & \\
<Line>: x=144, y=220 & x=207, y=220 with a sweep angle of 64 degrees & \\
<Arc>: x=128, y=204, $\alpha$=64, f=1 & and a counter-clockwise direction, a line to & \\
<Line>: x=128, y=128 & x=144, y=220, and an arc to x=128, y=204 with & \\
\texttt{<SOL>} & a sweep angle of 64 degrees and a counter-clockwise & \\
<Arc>: x=144, y=114, $\alpha$=64, f=1 & direction. Finally, draw a line to x=128, y=128. & \\
<Line>: x=207, y=114 & Repeat the same sequence starting with an arc with & \\
<Arc>: x=220, y=128, $\alpha$=64, f=1 & an end-point at x=144, y=114, a sweep angle of & \\
<Line>: x=220, y=190 & 64 degrees, and a counter-clockwise direction. Then, & \\
<Arc>: x=207, y=204, $\alpha$=64, f=1 & extrude the 2D sketch by 148 units in the direction & \\
<Line>: x=144, y=204 & specified by $\theta$=192, $\varphi$=64, $\gamma$=192, originating at & \\
<Arc>: x=131, y=190, $\alpha$=64, f=1 & px=105, py=121, pz=40, and scaling the profile & \\
<Line>: x=131, y=128 & by a factor of 46. The extrusion will extend & \\
<Extrude>: $\theta$=192, $\varphi$=64, $\gamma$=192, px=105, py=121, pz=40, s=46, e1=148, e2=128, b=NewBodyFeatureOperation, u=OneSideFeatureExtentType & for a distance of 128 units in the direction given by u=OneSideFeatureExtentType and will be treated as a NewBodyFeatureOperation. & \\
\bottomrule
\end{tabular}
\caption{The original CCS and the first generated descriptions.}
\label{tab:comparison}

\end{table*}

\begin{table*}[t]
\resizebox{\textwidth}{!}{
\centering
\renewcommand{\arraystretch}{1.2}
\scriptsize
\begin{tabular}{p{0.7cm}p{7cm}p{7cm}c}  
\toprule
\textbf{Round} & \textbf{Reflection Output} & \textbf{Optimization Output} & \textbf{$\text{LCS}_{\text{ratio}}$} \\
\midrule
1 & 
\begin{minipage}[t]{6.8cm}  
\raggedright  
\textbf{Generated vs. Real CCS Differences:} \\
1. Arc End Points: Generated CCS ends at \( x=223, y=130 \); real CCS at \( x=220, y=128 \). \\
2. Line End Points: Generated CCS ends at \( x=207, y=114 \); real CCS at \( x=207, y=112 \). \\
3. Extra Commands: Real CCS includes a line to \( x=220, y=190 \); generated CCS specifies \( x=223, y=204 \). \\
4. Arc Variation: Generated CCS ends an arc at \( x=207, y=222 \); real CCS at \( x=207, y=204 \). \\
5. Missing Variables: Generated CCS omits \( e_2 \) (128 in real CCS). \\[4pt]
\textbf{Problems in Parametric Description:} \\
1. Ambiguous Arcs: Incomplete parameters cause endpoint inconsistencies. \\
2. Unclear Pathing: Lacks instructions for connecting arcs and lines or closing loops. \\
3. Rotation Inconsistencies: Arc rotation directions are unclear. \\
4. Missing Line Dimensions: No explicit dimensions for connecting lines. \\
5. Undefined Parameters: \( e_2 \) is missing from the description. \\[4pt]
\textbf{Improvement Suggestions:} \\
1. Define arcs explicitly with endpoints, radii, and centers. \\
2. Clarify pathing, including connection rules and loop closures. \\
3. Standardize rotation directions. \\
4. Specify line dimensions clearly. \\
5. Define all variables, including \( e_2 \). \\
6. Add visual references to support the description.
\end{minipage} 
& 
\begin{minipage}[t]{6.8cm}  
\raggedright
First, draw an arc with an end-point at \( x=144, y=112 \), a sweep angle of 64 degrees, and a counter-clockwise direction. Then, draw a line to \( x=207, y=112 \), followed by an arc to \( x=220, y=128 \) with a sweep angle of 64 degrees and a counter-clockwise direction. Next, draw a line to \( x=220, y=190 \), an arc to \( x=207, y=204 \) with a sweep angle of 64 degrees and a counter-clockwise direction, a line to \( x=144, y=204 \), and an arc to \( x=131, y=190 \) with a sweep angle of 64 degrees and a counter-clockwise direction. Finally, draw a line to \( x=131, y=128 \).

Repeat the same sequence starting with an arc with an end-point at \( x=144, y=114 \), a sweep angle of 64 degrees, and a counter-clockwise direction. Then, draw a line to \( x=207, y=114 \), followed by an arc to \( x=220, y=128 \) with a sweep angle of 64 degrees and a counter-clockwise direction. Next, draw a line to \( x=220, y=190 \), an arc to \( x=207, y=204 \) with a sweep angle of 64 degrees and a counter-clockwise direction, a line to \( x=144, y=204 \), and an arc to \( x=131, y=190 \) with a sweep angle of 64 degrees and a counter-clockwise direction. Finally, draw a line to \( x=131, y=128 \).

Then, extrude the 2D sketch by 148 units in the direction specified by \( \theta=192 \), \( \phi=64 \), \( \gamma=192 \), originating at \( px=105 \), \( py=121 \), \( pz=40 \), and scaling the profile by a factor of 46. The extrusion will extend for a distance of 128 units in the direction given by u=OneSideFeatureExtentType and be treated as a NewBodyFeatureOperation.
\end{minipage} 
& 0.737 \\
\midrule
2 & 
\begin{minipage}[t]{6.8cm}  
\raggedright
\textbf{Generated vs. Real CCS Differences:} \\
1. Arc Direction: Generated CCS uses \( f=-1 \) (clockwise), real CCS uses \( f=1 \) (counter-clockwise). \\
2. Arc Endpoint Discrepancy: Generated CCS ends at \( x=220, y=128 \), real CCS at \( x=223, y=128 \). \\
3. Line Mismatch: Generated CCS connects to \( x=220, y=190 \), real CCS to \( x=223, y=204 \). \\
4. Arc Endpoint Discrepancy: Later arcs end at \( x=207, y=220 \), instead of \( x=207, y=204 \). \\
5. Fewer Commands: Generated CCS omits lines and arcs. \\
6. Extrusion Parameters: Generated CCS uses \( e_1=0 \) and omits \( e_2 \); real CCS uses \( e_1=148 \) and \( e_2=128 \). \\[4pt]
\textbf{Problems in CAD Parametric Description:} \\
1. Unclear Arc Directions: \( f \) not explicitly defined. \\
2. Ambiguous Arc Coordinates: Lack of precise endpoint relationships. \\
3. Endpoint Proximity Issues: Close arc endpoints cause confusion. \\
4. Incomplete Extrusion Definitions: \( e_1 \) and \( e_2 \) inadequately described. \\[4pt]
\textbf{Suggestions for Improvement:} \\
1. Define arc directions explicitly. \\
2. Derive arc endpoints explicitly from previous commands. \\
3. Implement validation checks for consistency. \\
4. Provide clear extrusion parameter definitions. \\
5. Ensure sequential coherence in parametric commands.
\end{minipage} 
& 
\begin{minipage}[t]{6.8cm}  
\raggedright
First, draw an arc with an end-point at \( x=144, y=112 \), a sweep angle of 64 degrees, and a counter-clockwise direction (\( f=1 \)). Then, draw a line to \( x=207, y=112 \). Next, draw an arc to \( x=223, y=128 \) with a sweep angle of 64 degrees and a counter-clockwise direction (\( f=1 \)). Draw a line to \( x=223, y=204 \). Then, draw an arc to \( x=207, y=220 \) with a sweep angle of 64 degrees and a counter-clockwise direction (\( f=1 \)). After that, draw a line to \( x=144, y=220 \). Then, draw an arc to \( x=128, y=204 \) with a sweep angle of 64 degrees and a counter-clockwise direction (\( f=1 \)). Finally, draw a line to \( x=128, y=128 \).

Repeat the same sequence starting with an arc with an end-point at \( x=144, y=114 \), a sweep angle of 64 degrees, and a counter-clockwise direction (\( f=1 \)). Then, draw a line to \( x=207, y=114 \). Next, draw an arc to \( x=220, y=128 \) with a sweep angle of 64 degrees and a counter-clockwise direction (\( f=1 \)). After that, draw a line to \( x=220, y=190 \). Then, draw an arc to \( x=207, y=204 \) with a sweep angle of 64 degrees and a counter-clockwise direction (\( f=1 \)). Draw a line to \( x=144, y=204 \). Next, draw an arc to \( x=131, y=190 \) with a sweep angle of 64 degrees and a counter-clockwise direction (\( f=1 \)). Finally, draw a line to \( x=131, y=128 \).

Then, extrude the 2D sketch by 148 units in the direction specified by \( \theta=192 \), \( \phi=64 \), \( \gamma=192 \), originating at \( px=105 \), \( py=121 \), \( pz=40 \), and scaling the profile by a factor of 46. The extrusion will extend for a distance of 128 units in the direction given by u=OneSideFeatureExtentType and will be treated as a NewBodyFeatureOperation with parameters \( e_1=148 \) and \( e_2=128 \).
\end{minipage} 
& 1.000 \\
\bottomrule
\end{tabular}
}
\caption{Multiple-round Reflection and Optimization Results for CCS Generation.}
\label{tab:reflection-results}
\end{table*}

\subsubsection{Appearance Description Annotation}

The examples of appearance description generation are shown in Table~\ref{tab:annotation_examples}. For each data sample, in the description generation stage, both multi-view descriptions (generated by the VLLM) and point cloud descriptions (generated by the point-cloud-based large language model) are produced simultaneously. The two descriptions are verified using a large language model in the description quality control stage. For example, in sample No. 00247561, the multi-view and point cloud descriptions were inconsistent, and the model's output was flagged as False. This sample was then sent for manual annotation. In contrast, for sample No. 00041425, the multi-view and point cloud descriptions were consistent, and the model's output was True, allowing for generating a unified description based on both representations.

\begin{table*}[h]

\scriptsize
\centering
\newcolumntype{P}[1]{>{\centering\arraybackslash}m{#1}}
\newcolumntype{M}[1]{>{\raggedright\arraybackslash}m{#1}}
\begin{tabular}{P{2.5cm}M{3cm}M{3cm}P{2.5cm}M{3cm}}
\toprule
\textbf{Model Visualization} & \textbf{Point Cloud Description} & \textbf{Multi-View Description} & \textbf{LLM Determination} & \textbf{Comprehensive Description} \\
\midrule
\vspace{0pt}\includegraphics[width=2.5cm]{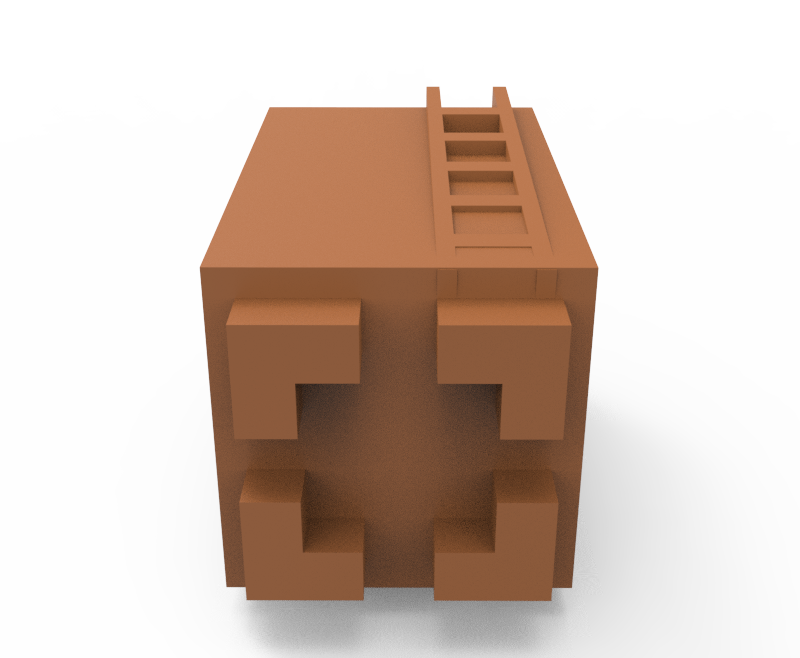}
& The point cloud represents a 3D model of an engine, exhibiting intricate details. It includes components such as cylinders, connectors, and automotive parts, though the specific engine type is not identified. The model is likely intended for educational or demonstration purposes in mechanics or engineering. 
& The object is cube-shaped with a ladder on its top surface. It consists of a large cube, four smaller L-shaped protrusions on one side, and a ladder structure. The geometric form is simple, with clearly defined edges and shapes, which could be a building block or part of a larger assembly. 
& FALSE (Manually Annotated) 
& This 3D object has a cubic form with a ladder attached to its upper surface. It consists of a main cube, four smaller L-shaped protrusions, and a ladder-like structure. \\
\midrule
\vspace{0pt}\includegraphics[width=2.5cm]{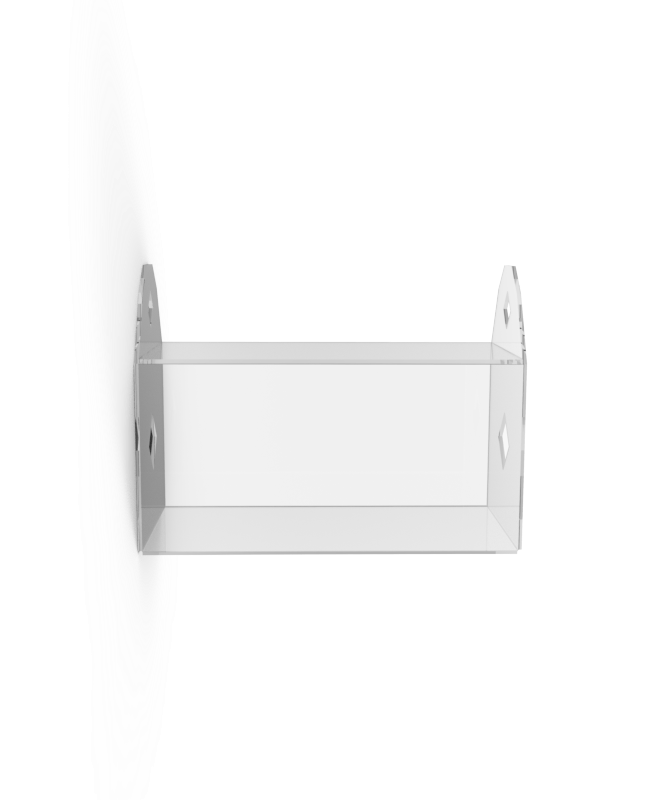}
& The 3D point cloud represents a square box with a closed and smooth surface, suggesting it could be a container or storage object. Its shape implies various uses, such as storing small items or being a building block. The lack of overall context makes its practical use unclear, but it may be intended for architectural modeling, digital art, or advanced mathematical modeling. 
& The 3D model appears as a cube with a cylindrical protrusion on one side, featuring a curved top and a flat bottom. The model has a cube base, a cylindrical body, and a curved top. The smooth surface and tapering cylindrical body suggest the model could serve as a container or vessel. 
& TRUE (Large Model Annotation) 
& The 3D model has a square box (cube) as the base. One side of the cube has a cylindrical protrusion with a curved top and flat bottom. The cylindrical body tapers subtly towards the curved top. \\
\bottomrule
\end{tabular}
\caption{Point Cloud and Multi-View Descriptions with Model Determinations and Comprehensive Analysis.}
\label{tab:annotation_examples}
\end{table*}

\subsubsection{LLMCAD Enhancement}

\begin{figure*}
    \centering
    \includegraphics[width=0.83\linewidth]{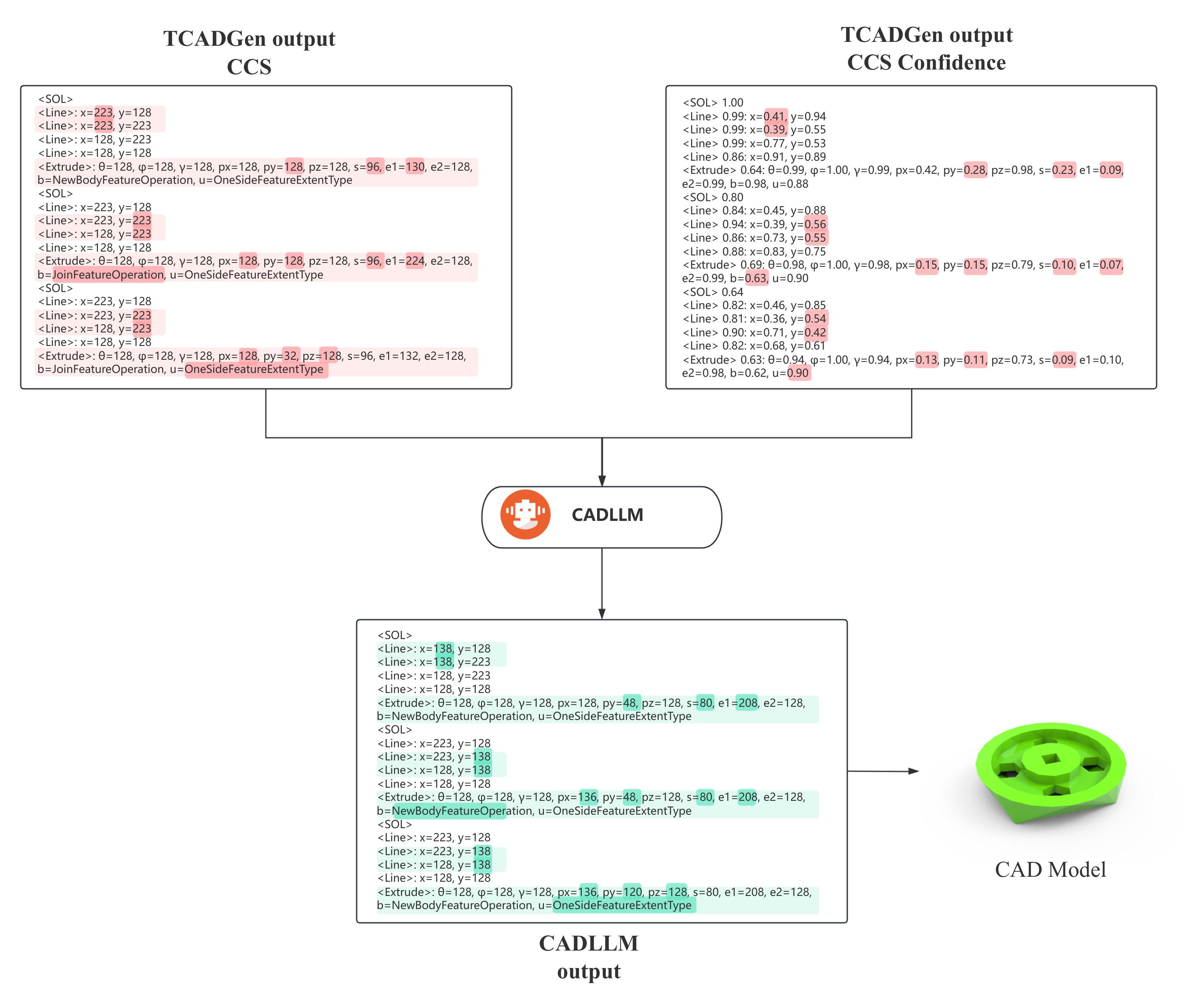}
    \caption{Final CCS generated by CAD LLM using TCADGen CCS output and confidence, with red and green areas indicating modified sections.}
    \label{fig:CADLLM_examp}
\end{figure*}

Figure~\ref{fig:CADLLM_examp} shows the final CCS generated by the CADLLM based on the TCADGen CCS output and confidence output. The red and green areas highlight the modified sections, which correspond to the parts where the confidence prediction was lower. This demonstrates that the CAD LLM task involves generating a new, correct CCS based on the existing parameter descriptions, considering the confidence levels and the TCADGen output. In this example, the CCS output from the CAD LLM exactly matches the ground truth.


\begin{table*}[htbp]
\centering
\resizebox{\textwidth}{!}{
\renewcommand{\arraystretch}{1.5}  
\setlength{\tabcolsep}{6pt}  
\begin{tabular}{l|ccccc}  
\toprule
\multirow{2}{*}{Models} & \multicolumn{5}{c}{Generated Results} \\
\cmidrule{2-6}
& 00001213 & 00004252 & 00005341 & 00012350 & 00016271 \\
\midrule
DeepCAD & \adjustbox{valign=c}{\includegraphics[width=2.2cm,height=2.2cm,keepaspectratio]{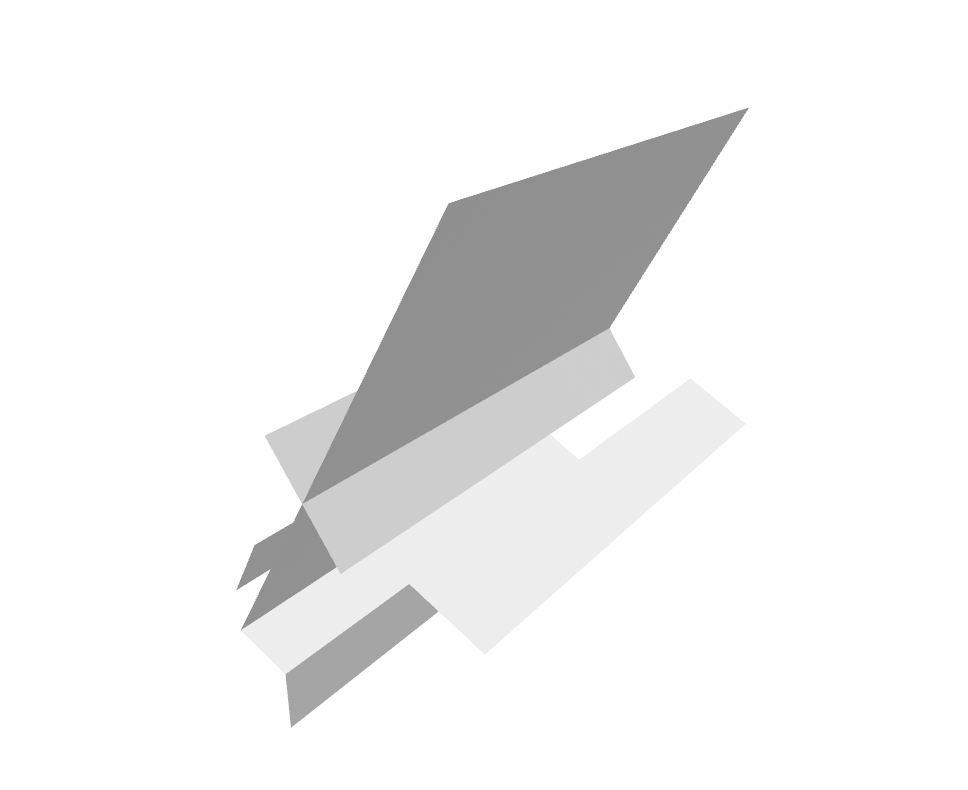}} & \adjustbox{valign=c}{\includegraphics[width=2.2cm,height=2.2cm,keepaspectratio]{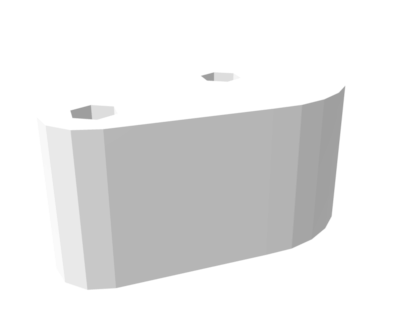}} & \adjustbox{valign=c}{\includegraphics[width=2.2cm,height=2.2cm,keepaspectratio]{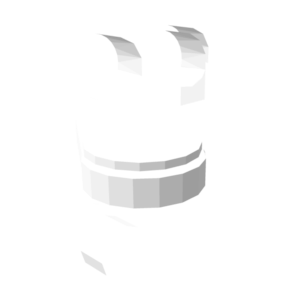}} & \adjustbox{valign=c}{\includegraphics[width=2.2cm,height=2.2cm,keepaspectratio]{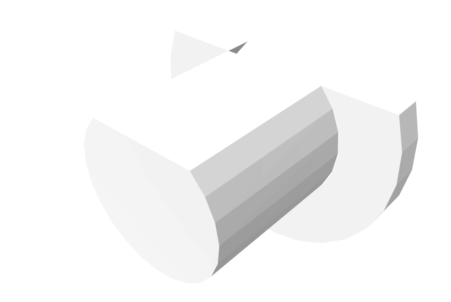}} & \adjustbox{valign=c}{\includegraphics[width=2.2cm,height=2.2cm,keepaspectratio]{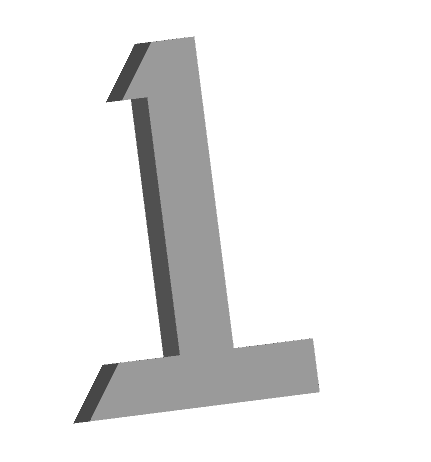}} \\
\midrule
Text2CAD & \adjustbox{valign=c}{\includegraphics[width=2.2cm,height=2.2cm,keepaspectratio]{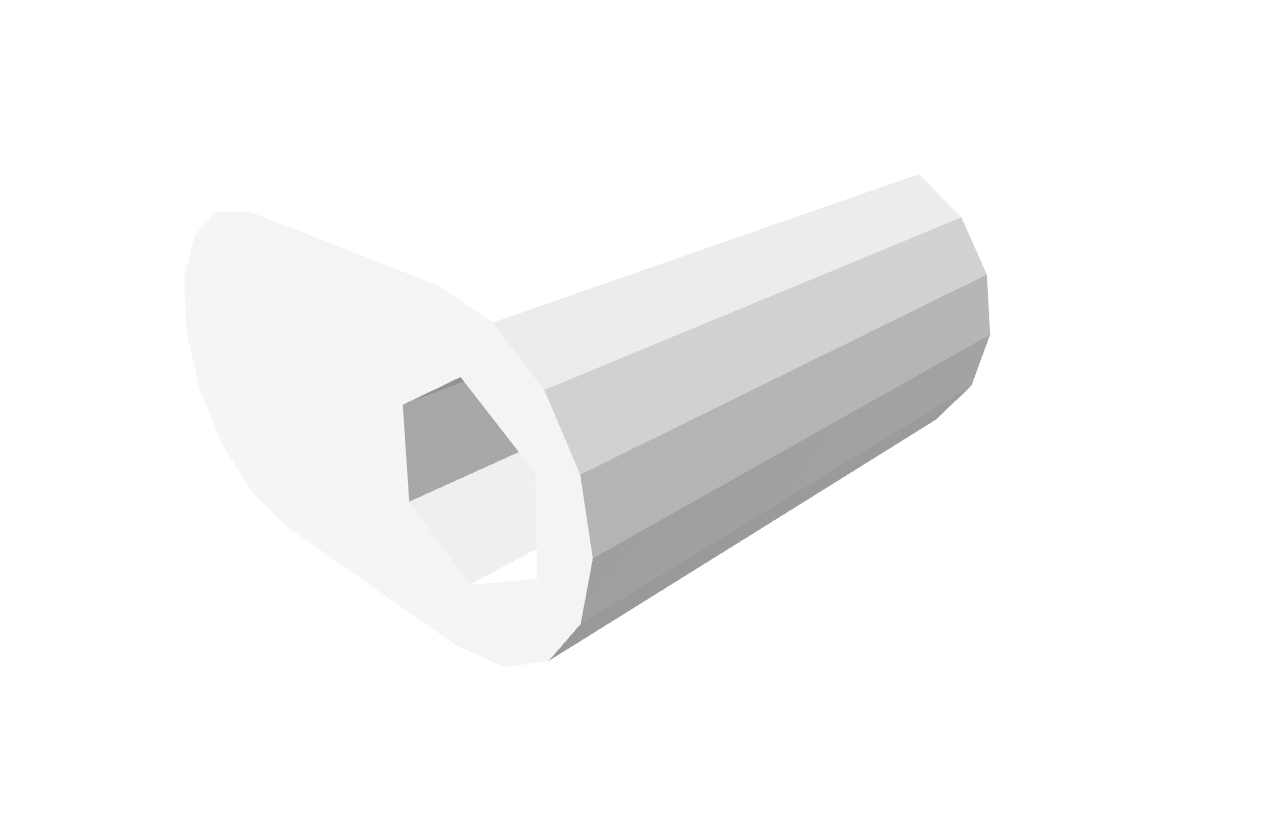}} & \adjustbox{valign=c}{\includegraphics[width=2.2cm,height=2.2cm,keepaspectratio]{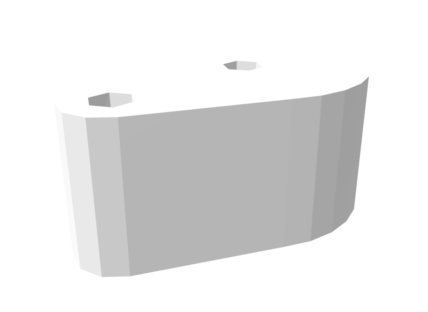}} & \adjustbox{valign=c}{\includegraphics[width=2.2cm,height=2.2cm,keepaspectratio]{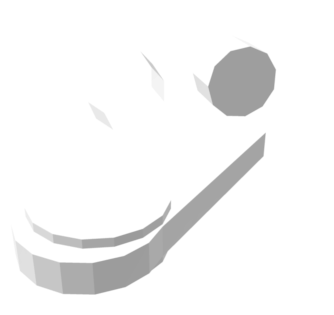}} & \adjustbox{valign=c}{\includegraphics[width=2.2cm,height=2.2cm,keepaspectratio]{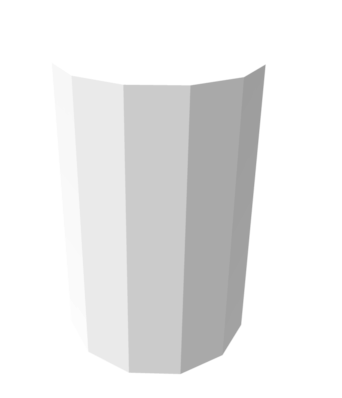}} & \adjustbox{valign=c}{\includegraphics[width=2.2cm,height=2.2cm,keepaspectratio]{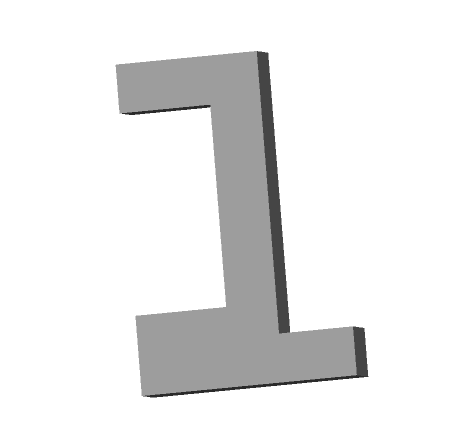}} \\
\midrule
TCADGen+CADLLM & \adjustbox{valign=c}{\includegraphics[width=2.2cm,height=2.2cm,keepaspectratio]{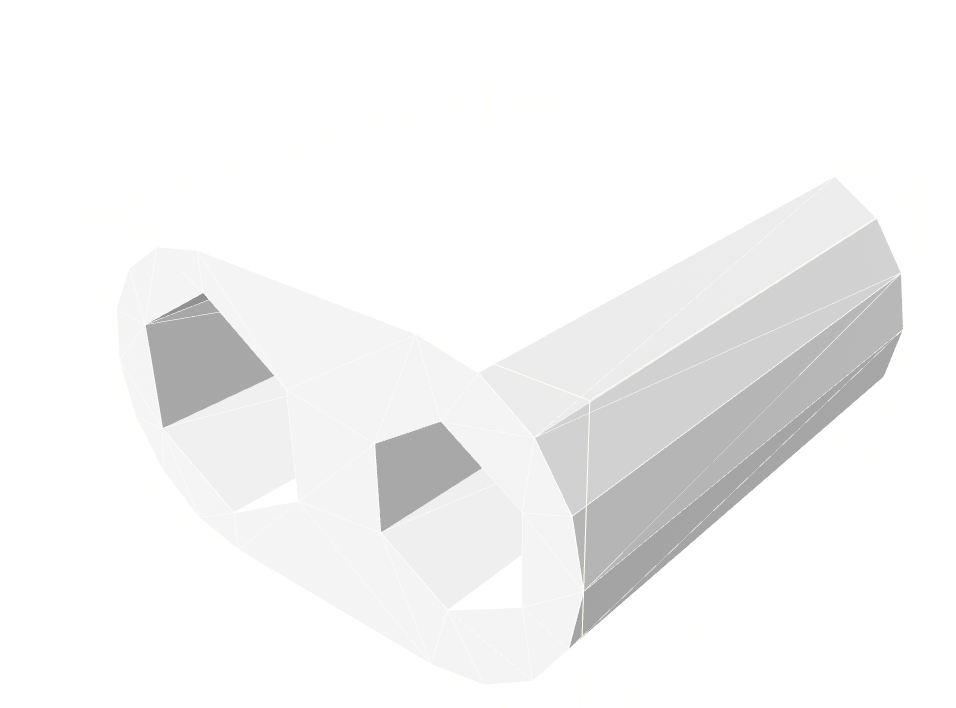}} & \adjustbox{valign=c}{\includegraphics[width=2.2cm,height=2.2cm,keepaspectratio]{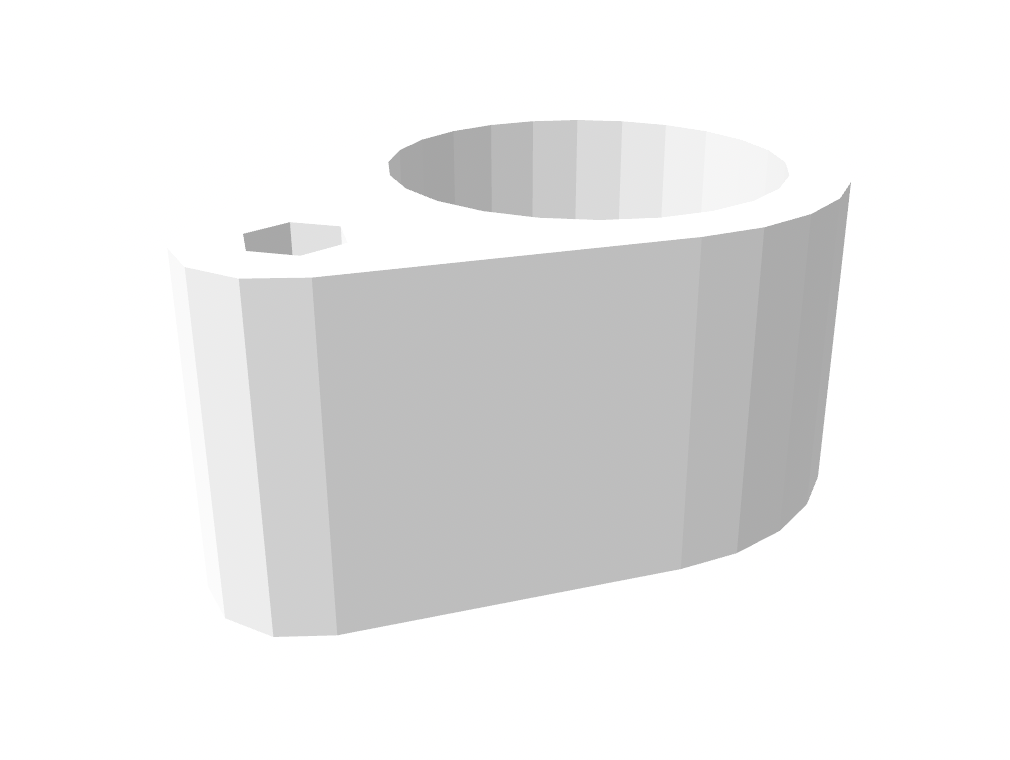}} & \adjustbox{valign=c}{\includegraphics[width=2.2cm,height=2.2cm,keepaspectratio]{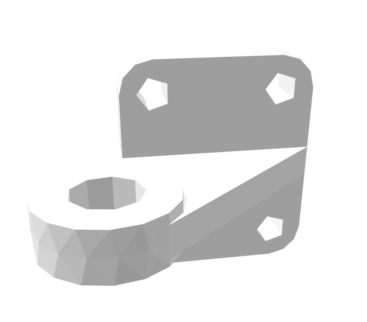}} & \adjustbox{valign=c}{\includegraphics[width=2.2cm,height=2.2cm,keepaspectratio]{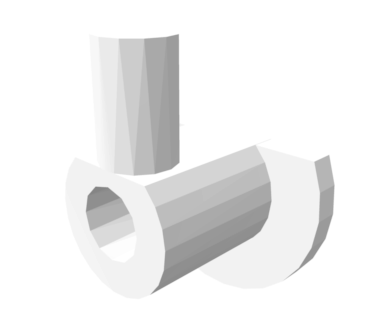}} & \adjustbox{valign=c}{\includegraphics[width=2.2cm,height=2.2cm,keepaspectratio]{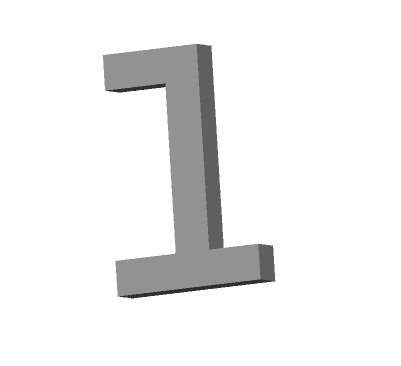}} \\
\midrule
Ground Truth & \adjustbox{valign=c}{\includegraphics[width=2.2cm,height=2.2cm,keepaspectratio]{image/Appendix/show_difference/00001213_gt}} & \adjustbox{valign=c}{\includegraphics[width=2.2cm,height=2.2cm,keepaspectratio]{image/Appendix/show_difference/00004252_gt}} & \adjustbox{valign=c}{\includegraphics[width=2.2cm,height=2.2cm,keepaspectratio]{image/Appendix/show_difference/00005341_gt}} & \adjustbox{valign=c}{\includegraphics[width=2.2cm,height=2.2cm,keepaspectratio]{image/Appendix/show_difference/00012350_gt}} & \adjustbox{valign=c}{\includegraphics[width=2.2cm,height=2.2cm,keepaspectratio]{image/Appendix/show_difference/00016271_gt}} \\
\bottomrule
\end{tabular}
}
\caption{CAD model effects generated by different methods.}
\label{tab:cad_comparison}
\end{table*}

\subsection{The Ability of LLMs to Generate Parameter Descriptions}\label{sec:LLMs to Generate Parameter Descriptions}

To assess the ability of LLMs in generating CAD modeling descriptions, we introduce a two-phase evaluation framework consisting of forward generation and backward validation. We evaluate a test set of $1000$ samples that reflect the natural distribution of our dataset. LLMs generate CCS descriptions based on given prompts in the forward generation phase. The backward validation phase employs different LLMs to verify these generated descriptions.

We define four evaluation criteria based on $\text{LCS}_{\text{ratio}}$:

\begin{enumerate}
    \item Exact matches: $\text{LCS}_{\text{ratio}} = 1$
    \item High-quality matches: $\text{LCS}_{\text{ratio}} \geq 0.98$
    \item Acceptable matches: $\text{LCS}_{\text{ratio}} \geq 0.9$
    \item Mean $\text{LCS}_{\text{ratio}}$ across all test samples
\end{enumerate}

The proportions for each criterion are computed as follows:

\begin{align}
\text{Proportion}_{\text{criterion}} = \frac{\text{count}(\text{LCS}_{\text{ratio}} \geq \text{criterion})}{N},
\end{align}

where $N$ is the total number of samples, and $\text{LCS}_{\text{ratio}}^{(i)}$ is the $\text{LCS}_{\text{ratio}}$ for the $i$-th sample.The average $\text{LCS}_{\text{ratio}}$ is calculated as:

\begin{align}
\text{Average LCS}_{\text{ratio}} = \frac{1}{N} \sum_{i=1}^{N} \text{LCS}_{\text{ratio}}^{(i)},
\end{align}

We present the raw evaluation data in Table~\ref{tab:lcs_raw}. Table~\ref{tab:lcs_proportion} summarizes the results for the Best Reverse Validation Model, where the model with the highest proportion of LCS Greater than 0.9 for each forward generation model is selected as the Best Reverse Validation Model. The results show strong performance from both closed-source and open-source models. In forward generation, Gemini-1.5-Pro and GPT-4o achieve $\text{LCS}{\text{ratio}}$ values above 93.9\%. Among open-source models, Meta-Llama-3.1-70B-Instruct-Turbo and Meta-Llama-3.1-405B-Instruct-Turbo perform well. Notably, Gemini-2-27B-IT shows competitive results despite its smaller size. For backward validation, we define the Best Reverse Model as the one that most frequently achieves $\text{LCS}{\text{ratio}} > 0.9$. Gemini-2-27B-IT is the most reliable model for this validation task, demonstrating its effectiveness in accurately verifying CCS descriptions.

\begin{table*}[b]
\centering
\small
\setlength{\tabcolsep}{4.2pt}  

\label{tab:RQ3-data-size-impact}
\begin{tabular}{l|ccc|c|cccc|cc}
\hline
\textbf{Data} & \multicolumn{3}{c|}{\textbf{Command Prediction}} & \textbf{Param} & \multicolumn{4}{c|}{\textbf{Command-Type Performance}} & \multicolumn{2}{c}{\textbf{CCS}} \\
\cline{2-11}
\textbf{Size} & ACC & F1 & AUC & \textbf{ACC} & Line & Arc & Circle & Extrude & Accuracy & Avg $\text{LCS}_{\text{ratio}}$ \\
\hline
0 & 0.532 & 0.769 & 0.851 & 0.566 & 0.939 & 0.564 & 0.756 & 0.675 & 0.160 & 0.622 \\
100 & 0.849 & 0.818 & 0.871 & 0.924 & 0.928 & 0.678 & 0.814 & 0.788 & 0.567 & 0.912 \\
500 & 0.932 & 0.907 & 0.934 & 0.970 & 0.931 & 0.855 & 0.907 & 0.900 & 0.789 & 0.972 \\
1000 & \textbf{0.966} & \textbf{0.947} & \textbf{0.962} & \textbf{0.983} & \textbf{0.979} & \textbf{0.924} & \textbf{0.960} & \textbf{0.946} & \textbf{0.864} & \textbf{0.983} \\
\hline
\end{tabular}
\caption{Performance analysis across different training data sizes.}
\label{tab:RQ3-data}
\end{table*}

\begin{table*}[htbp]
    \centering

    \large
    \renewcommand{\arraystretch}{1}
    \begin{adjustbox}{max width=\textwidth, max height=\textheight}
    \begin{tabular}{lcccccccc}
        \toprule
        \diagbox[width=7cm]{Forward Generation Models}{Reverse Verification Models}
        & \textbf{Similarity}
        & \textbf{GPT-4o-mini}
        & \textbf{Gemini-1.5-Flash}
        & \textbf{Google\_Gemma-2-27B-it}
        & \textbf{Google\_Gemma-2-9B-it}
        & \textbf{Meta-Llama-3.1-70B}
        & \textbf{Meta-Llama-3.1-405B}
        & \textbf{Qwen2.5-72B} \\
        \midrule
        \textbf{claude-3-5-haiku-20241022} 
        & 1 & 0.2 & 0.16 & 0.36 & 0.08 & 0.24 & 0.24 & 0.24 \\
        & 0.98 & 0.44 & 0.36 & 0.52 & 0.36 & 0.52 & 0.56 & 0.52 \\
        & 0.9 & 0.68 & 0.64 & 0.80 & 0.56 & 0.84 & 0.76 & 0.80 \\
        & \textbf{Average} & 0.905 & 0.866 & 0.932 & 0.829 & 0.927 & 0.938 & 0.929 \\
        \midrule
        \textbf{claude-3-5-sonnet-20241022}
        & 1 & 0.24 & 0.24 & 0.32 & 0.12 & 0.32 & 0.24 & 0.24 \\
        & 0.98 & 0.52 & 0.76 & 0.84 & 0.48 & 0.68 & 0.80 & 0.80 \\
        & 0.9 & 0.76 & 0.84 & 0.88 & 0.68 & 0.80 & 0.84 & 0.88 \\
        & \textbf{Average} & 0.896 & 0.939 & 0.942 & 0.841 & 0.937 & 0.932 & 0.943 \\
        \midrule
        \textbf{ERNIE-4.0-8K-Latest}
        & 1 & 0.385 & 0.346 & 0.346 & 0.038 & 0.346 & 0.385 & 0.346 \\
        & 0.98 & 0.577 & 0.615 & 0.538 & 0.346 & 0.577 & 0.615 & 0.615 \\
        & 0.9 & 0.769 & 0.846 & 0.731 & 0.731 & 0.846 & 0.808 & 0.846 \\
        & \textbf{Average} & 0.9 & 0.918 & 0.915 & 0.877 & 0.929 & 0.935 & 0.933 \\
        \midrule
        \textbf{ERNIE-Speed-8K}
        & 1 & 0.141 & 0.141 & 0.051 & 0.0 & 0.141 & 0.15 & 0.141 \\
        & 0.98 & 0.224 & 0.229 & 0.103 & 0.206 & 0.235 & 0.245 & 0.235 \\
        & 0.9 & 0.418 & 0.424 & 0.462 & 0.412 & 0.412 & 0.449 & 0.418 \\
        & \textbf{Average} & 0.661 & 0.669 & 0.672 & 0.632 & 0.67 & 0.684 & 0.669 \\
        \midrule
        \textbf{gemini-1.5-flash}
        & 1 & 0.373 & 0.363 & 0.43 & 0.13 & 0.397 & 0.404 & 0.417 \\
        & 0.98 & 0.573 & 0.633 & 0.717 & 0.53 & 0.627 & 0.68 & 0.65 \\
        & 0.9 & 0.817 & 0.807 & 0.883 & 0.763 & 0.843 & 0.859 & 0.853 \\
        & \textbf{Average} & 0.929 & 0.937 & 0.95 & 0.905 & 0.939 & 0.945 & 0.944 \\
        \midrule
        \textbf{gemini-1.5-pro}
        & 1 & 0.333 & 0.45 & 0.487 & 0.067 & 0.453 & 0.429 & 0.4 \\
        & 0.98 & 0.56 & 0.743 & 0.807 & 0.472 & 0.71 & 0.751 & 0.687 \\
        & 0.9 & 0.867 & 0.883 & 0.947 & 0.829 & 0.887 & 0.87 & 0.92 \\
        & \textbf{Average} & 0.942 & 0.956 & 0.973 & 0.928 & 0.955 & 0.953 & 0.964 \\
        \midrule
        \textbf{gemma-2-27b-it}
        & 1 & 0.217 & 0.19 & 0.253 & 0.027 & 0.27 & 0.282 & 0.2 \\
        & 0.98 & 0.413 & 0.463 & 0.563 & 0.387 & 0.487 & 0.533 & 0.43 \\
        & 0.9 & 0.787 & 0.787 & 0.877 & 0.78 & 0.82 & 0.861 & 0.823 \\
        & \textbf{Average} & 0.912 & 0.923 & 0.947 & 0.9 & 0.929 & 0.948 & 0.932 \\
        \midrule
        \textbf{gemma-2-9b-it}
        & 1 & 0.397 & 0.387 & 0.407 & 0.053 & 0.4 & 0.41 & 0.415 \\
        & 0.98 & 0.647 & 0.667 & 0.677 & 0.62 & 0.687 & 0.68 & 0.682 \\
        & 0.9 & 0.78 & 0.8 & 0.797 & 0.773 & 0.787 & 0.782 & 0.793 \\
        & \textbf{Average} & 0.902 & 0.91 & 0.914 & 0.898 & 0.908 & 0.911 & 0.909 \\
        \midrule
        \textbf{gpt-4o}
        & 1 & 0.435 & 0.421 & 0.462 & 0.1 & 0.472 & 0.43 & 0.442 \\
        & 0.98 & 0.736 & 0.773 & 0.773 & 0.589 & 0.783 & 0.795 & 0.757 \\
        & 0.9 & 0.87 & 0.916 & 0.9 & 0.86 & 0.9 & 0.93 & 0.899 \\
        & \textbf{Average} & 0.951 & 0.964 & 0.962 & 0.939 & 0.963 & 0.971 & 0.957 \\
        \midrule
        \textbf{gpt-4o-mini}
        & 1 & 0.453 & 0.403 & 0.44 & 0.123 & 0.463 & 0.452 & 0.477 \\
        & 0.98 & 0.677 & 0.697 & 0.71 & 0.667 & 0.723 & 0.756 & 0.707 \\
        & 0.9 & 0.813 & 0.837 & 0.857 & 0.82 & 0.867 & 0.87 & 0.86 \\
        & \textbf{Average} & 0.933 & 0.946 & 0.949 & 0.932 & 0.944 & 0.954 & 0.946 \\
        \midrule
        \textbf{Llama-3.1-405B-Instruct}& 1 & 0.457 & 0.496 & 0.558 & 0.068 & 0.568 & 0.517 & 0.565 \\
        & 0.98 & 0.662 & 0.766 & 0.856 & 0.522 & 0.827 & 0.779 & 0.809 \\
        & 0.9 & 0.856 & 0.874 & 0.939 & 0.802 & 0.928 & 0.858 & 0.921 \\
        & \textbf{Average} & 0.943 & 0.967 & 0.976 & 0.923 & 0.968 & 0.959 & 0.971 \\
        \midrule
        \textbf{Llama-3.1-70B-Instruct}& 1 & 0.534 & 0.605 & 0.611 & 0.071 & 0.639 & 0.605 & 0.642 \\
        & 0.98 & 0.726 & 0.838 & 0.841 & 0.541 & 0.845 & 0.855 & 0.831 \\
        & 0.9 & 0.902 & 0.943 & 0.946 & 0.841 & 0.912 & 0.924 & 0.936 \\
        & \textbf{Average} & 0.963 & 0.979 & 0.981 & 0.941 & 0.972 & 0.969 & 0.978 \\
        \midrule
        \textbf{Qwen2.5-72B-Instruct-Turbo}
        & 1 & 0.25 & 0.31 & 0.323 & 0.017 & 0.363 & 0.283 & 0.29 \\
        & 0.98 & 0.45 & 0.59 & 0.613 & 0.33 & 0.637 & 0.595 & 0.573 \\
        & 0.9 & 0.78 & 0.823 & 0.817 & 0.687 & 0.85 & 0.848 & 0.81 \\
        & \textbf{Average} & 0.912 & 0.936 & 0.942 & 0.883 & 0.947 & 0.946 & 0.935 \\
        \bottomrule
    \end{tabular}
    \end{adjustbox}
    \caption{Proportions of results generated by the forward generation model with $\text{LCS}_{\text{ratio}}$ values of 1, 0.98, and 0.9, as well as the average $\text{LCS}_{\text{ratio}}$ for the backward validation model, which verifies the generated CCS descriptions.}
    \label{tab:lcs_raw}
\end{table*}

\begin{table*}[htbp]
\centering

\resizebox{\textwidth}{!}{
\begin{tabular}{l l c}
\toprule
\textbf{Forward Model} & \textbf{Best Reverse Model} & \textbf{Proportion of LCS Greater than 0.9} \\
\midrule
gemini-1.5-pro & gemma-2-27b-it& 0.947 \\
Llama-3.1-70B-Instruct-Turbo& gemma-2-27b-it& 0.946 \\
Llama-3.1-405B-Instruct-Turbo& gemma-2-27b-it& 0.939 \\
gpt-4o & Llama-3.1-405B-Instruct& 0.930 \\
ERNIE-3.5-8K & gemini-1.5-flash & 0.906 \\
gemini-1.5-flash & gemma-2-27b-it& 0.883 \\
claude-3-5-sonnet-20241022 & gemma-2-27b-it& 0.880 \\
gemma-2-27b-it & gemma-2-27b-it& 0.877 \\
gpt-4o-mini & Llama-3.1-405B-Instruct& 0.870 \\
Qwen2.5-72B-Instruct-Turbo & Llama-3.1-70B-Instruct& 0.850 \\
ERNIE-4.0-8K-Latest & gemini-1.5-flash & 0.846 \\
claude-3-5-haiku-20241022 & Llama-3.1-70B-Instruct& 0.840 \\
ERNIE-4.0-Turbo-8K & gemma-2-27b-it& 0.827 \\
gemma-2-9b-it & gemini-1.5-flash & 0.800 \\
ERNIE-Speed-8K & gemma-2-27b-it& 0.462 \\
\bottomrule
\end{tabular}
}
\caption{Proportion of LCS Greater than 0.9 for Different Forward and Reverse Models in the Two-Phase Evaluation Framework.}
\label{tab:lcs_proportion}
\end{table*}

\subsection{Prompts Used in Experiments}
Table~\ref{tab:process-prompts} presents all the prompt templates used in this experiment.


\clearpage
\begin{onecolumn}
\begin{xltabular}{\textwidth}{>{\centering\arraybackslash}p{2. cm} X}

\caption{Prompts Used in Experiments \label{tab:process-prompts}}\\

\toprule
\textbf{Process Name} & \textbf{Prompt} \\
\midrule
\endfirsthead

\multicolumn{2}{c}%
{{\bfseries Table \thetable{} -- continued from previous page}} \\
\toprule
\textbf{Process Name} & \textbf{Prompt} \\
\midrule
\endhead

\midrule
\multicolumn{2}{r}{{Continued on next page}} \\
\endfoot

\bottomrule
\endlastfoot

Multi-view analysis & You are an experienced CAD engineer tasked with providing natural language descriptions of design objects based on images. Your goal is to create clear, specific descriptions for junior designers to understand and model, focusing on shape features, structural elements, and spatial relationships.

You will be given an image. Your task is to analyze this description and generate four concise sentences that describe:

\begin{enumerate}
    \item What the 3D model looks like
    \item What it is composed of
    \item Its appearance
    \item What it can do or its purpose
\end{enumerate}
Follow these guidelines when generating your description:
\begin{itemize}
    \item Focus on the most important and distinctive features of the object
    \item Use clear and specific language to describe shapes, structures, and spatial relationships
    \item Avoid using phrases like ``this multi-view image shows'' or ``distinct views'' or explaining the image itself
    \item Do not include any explanations or additional commentary
    \item Do not describe the color, metallic sheen, or use words like ``blue'', ``shadow'', ``transparent'', ``metal'', ``plastic'', ``image'', ``black'', ``grey'', ``CAD model'', ``abstract'', ``orange'', ``purple'', ``golden'', ``smooth'', or ``green''
\end{itemize}
Present your four sentences in the following format:
\begin{itemize}
    \item Sentence describing what the 3D model looks like
    \item Sentence describing what it is composed of
    \item Sentence describing its appearance
    \item Sentence describing what it can do or its purpose
\end{itemize}
Ensure that each sentence is concise, clear, and adheres to the abovementioned guidelines. \\
\midrule
Point cloud analysis & You will be analyzing a 3D point cloud to describe its geometric structure. Your task is to generate four precise sentences, each focusing on a different aspect of the object represented by the point cloud:

\begin{enumerate}
    \item Overall Shape
    \item Components
    \item Structural Details
    \item Function
\end{enumerate}

For each aspect, follow these specific guidelines:

\begin{itemize}
    \item \textbf{Overall Shape:} Describe the basic geometric form and approximate dimensions. Focus on the general shape and proportions.
    \item \textbf{Components:} List the main structural elements and their spatial relationships. Identify distinct parts and how they connect or relate to each other.
    \item \textbf{Structural Details:} Describe key geometric features, patterns, and surface characteristics. Focus on form rather than color or texture.
    \item \textbf{Function:} State the most likely practical purpose based purely on the form and structure. Make an educated guess about the object's intended use.
\end{itemize}

Important guidelines:

\begin{itemize}
    \item Be specific and detailed in your descriptions.
    \item Avoid any mention of colors, materials, or textures that cannot be definitively determined from point cloud data alone.
    \item Focus solely on geometry, form, and function.
    \item Use precise language and, where appropriate, include approximate measurements or proportions.
\end{itemize}

Keep descriptions focused on geometry, form, and function. Avoid any references to color, material, or texture that cannot be definitively determined from point cloud data alone. Write your analysis in four separate sentences, one for each aspect. Do not label or number the sentences. Ensure that each sentence flows naturally into the next.\\
\midrule
Appearance description generation and verification & 
You are a parameter in analyzing and describing 3D models based on point cloud data and multiple perspective descriptions. Your task is to analyze the compatibility between two descriptions and provide a synthesized description when they share fundamental geometric characteristics.

You will be given two input descriptions:

\begin{itemize}
    \item \texttt{\{POINT\_CLOUD\_DESCRIPTION\}}
    \item \texttt{\{MULTIPLE\_PERSPECTIVES\_DESCRIPTION\}}
\end{itemize}

Analyze these descriptions according to the following criteria:

\begin{enumerate}
    \item \textbf{Core Analysis Approach:}
    \begin{itemize}
        \item Extract basic geometric forms (shapes, volumes, structures)
        \item Look for ANY potential geometric compatibility
        \item Focus on finding similarities rather than differences
        \item Consider different ways of describing the same geometric concept
        \item Default to finding compatibility unless clearly contradictory
    \end{itemize}

    \item \textbf{Ignore Non-Essential Elements:}
    \begin{itemize}
        \item Color and material properties
        \item Surface textures and finishes
        \item Intended purpose or function
        \item Aesthetic qualities
        \item Subjective interpretations
    \end{itemize}
\end{enumerate}






 \\
\midrule
Parameter description generation & 
You are a parameter in CAD command descriptions. Your task is to describe a provided CAD command sequence based on specific requirements. Follow these instructions carefully:

First, familiarize yourself with these key terms and commands:
\begin{itemize}
    \item \textbf{〈SOL〉:} Denotes the start of a 2D closed curve. All commands prior to the Extrude command belong to this 2D curve.
    \item \textbf{L (Line):} x, y—Coordinates of the line's end-point, defining direction and length in the plane.
    \item \textbf{A (Arc):} x, y—Coordinates of the arc's end-point; $\alpha$—Sweep angle, indicating arc curvature; f—Counter-clockwise flag, specifying arc direction.
    \item \textbf{R (Circle):} x, y—Center coordinates of the circle; r—Radius, specifying circle size.
    \item \textbf{E (Extrude):} $\theta$, $\phi$, $\gamma$—Orientation of the sketch plane, defining its rotation and direction; px, py, pz—Origin of the sketch plane in 3D space; s—Scale factor, adjusting profile size; e1, e2—Extrude distances; b—Boolean operation type; u—Extrude direction.
\end{itemize}

Now, describe this CAD command sequence following these guidelines:
\begin{enumerate}
    \item Only output command descriptions—no additional content or explanations.
    \item Include each command, its parameters, and reflect the order of execution in your description.
    \item Describe how the Extrude command and its parameters transform the 2D curve into a 3D model.
    \item For the Extrude command, fully output the parameters b and u without changing them. Parameter b can be ``NewBodyFeatureOperation'', ``JoinFeatureOperation'', ``CutFeatureOperation'', or ``IntersectFeatureOperation''. Parameter u can be ``OneSideFeatureExtentType'', ``SymmetricFeatureExtentType'', or ``TwoSidesFThe parameterentType''.
    \item Do not change any parameter values in your description.
    \item Form a single, cohesive paragraph of CAD modeling guidance. Describe the sequence in natural language, including every parameter without omission.
\end{enumerate}

Here is the CAD command sequence you need to describe:

\begin{itemize}
    \item \texttt{\{CAD\_COMMAND\_SEQUENCE\}}
\end{itemize}

Your output should be a flowing, descriptive paragraph that guides the reader through the CAD modeling process, detailing each step and parameter in the order they appear in the command sequence. Do not use bullet points or numbered lists. Ensure your description is comprehensive, covering all aspects of the provided sequence.\\
\midrule

Parameter description verification & 
You are tasked with converting a CAD command description into a precise CAD operation sequence. Follow these instructions carefully:

\begin{enumerate}
    \item \textbf{Read the description carefully} and identify the CAD operations mentioned. These may include creating lines, arcs, circles, and performing extrude operations.
    \item \textbf{Convert each identified operation into the corresponding CAD command format} as follows:
    \begin{itemize}
        \item To start a new sketch:
        \begin{itemize}
            \item \texttt{<SOL>}
        \end{itemize}
        \item For a line:
        \begin{itemize}
            \item \texttt{<Line>}: x=<xValue>, y=<yValue>
        \end{itemize}
        \item For an arc:
        \begin{itemize}
            \item \texttt{<Arc>}: x=<xValue>, y=<yValue>, $\alpha$=<alphaValue>, f=<fValue>
        \end{itemize}
        \item For a circle:
        \begin{itemize}
            \item \texttt{<Circle>}: x=<xValue>, y=<yValue>, r=<radiusValue>
        \end{itemize}
        \item For an extrude operation:
        \begin{itemize}
            \item \texttt{<Extrude>}: $\theta$=<thetaValue>, $\varphi$=<phiValue>, $\gamma$=<gammaValue>, px=<pxValue>, py=<pyValue>, pz=<pzValue>, s=<sValue>, e1=<e1Value>, e2=<e2Value>, b=<extrudeOperation>, u=<extentType>
        \end{itemize}
    \end{itemize}
    \item \textbf{Arrange the converted operations in the sequence they appear in the description}. Start each new sketch with \texttt{<SOL>}.
    \item After converting all operations, end the CAD operation sequence with:
    \begin{itemize}
        \item \texttt{<EOS>} 
    \end{itemize}
    \item Ensure that you only output the CAD operation sequence without any additional explanations or comments.
    \item If any information is missing or unclear in the description, use reasonable default values or omit the parameter.
    \item Remember:
    \begin{itemize}
        \item There should be only one \texttt{<SOL>} tag per sketch.
        \item \texttt{<EOS>}  marks the end of the entire CAD operation sequence.
        \item Do not include any text or explanations outside of the specified command formats.
    \end{itemize}
    \item You will be given a CAD command description in the following format:
\begin{itemize}
    \item \texttt{\{CAD\_COMMAND\_DESCRIPTION\}}
\end{itemize}
    Output the resulting CAD operation sequence exactly as specified, with no additional commentary.
\end{enumerate}\\

\midrule

Reflection on parameter description issues & 
You are a CAD parametric description analysis parameter. Your task is to analyze the differences between a generated CAD Command Sequence (CCS) and a real CCS, identify problems in the CAD parametric description, and provide suggestions for improvement.
1. CAD parametric description:

\begin{itemize}
    \item \texttt{\{CAD\_DESCRIPTION\}}
\end{itemize}

2. Generated CCS based on the CAD description:

\begin{itemize}
    \item \texttt{\{GENERATED\_CCS\}}
\end{itemize}

\textbf{1. Differences between generated and real CCS:}
\begin{itemize}
    \item [\textbullet] Missing commands in the generated CCS: [List any specific commands present in the real CCS but missing in the generated one.]
    \item [\textbullet] Extra commands in the generated CCS: [List any commands that appear in the generated CCS but are not in the real CCS.]
    \item [\textbullet] Differences in command parameters or order: [List and describe any differences in parameters or the order of commands.]
\end{itemize}

\textbf{2. Problems in the CAD parametric description:}
\begin{itemize}
    \item [\textbullet] Missing information: [Discuss any information that is not clearly stated in the CAD parametric description, which may have led to missing or t commands.]
    \item [\textbullet] Incorrect specifications: [Identify any specifications that are inaccurate or potentially lead to errors in the generated CCS.]
    \item [\textbullet]inaccurate specifications [Highlight any parts of the CAD description that are ambiguous and could be interpreted in multiple ways.
\end{itemize}

\textbf{3. Improvement suggestions:}
\begin{itemize}
    \item [\textbullet] Clarify any ambiguous descriptions by providing more specific and detailed parameters.
    \item [\textbullet] Ensure that all necessary information for each command is provided to avoid missing commands or incomplete sequences.
    \item [\textbullet] Double-check the order of commands in the CAD description to match the expected sequence for correct execution.
    \item [\textbullet] Review the parameter specifications to ensure they are accurate and precise, avoiding potential discrepancies.
\end{itemize}

\\

\midrule

Parameter Description Regeneration &
You are a CAD parametric description modification parameter. Your task is to analyze and modify a given CAD parametric description to make it more accurate, complete, and unambiguous based on a real CCS sequence and a provided opinion. Follow these steps:

\begin{enumerate}
    \item \textbf{Review the provided CAD parametric description:}
    \begin{itemize}
        \item \texttt{\{CAD\_DESCRIPTION\}}
    \end{itemize}
    
    \item \textbf{Examine the real CCS sequence:}
    \begin{itemize}
        \item \texttt{\{REAL\_CCS\}}
    \end{itemize}

    \item \textbf{Consider the provided opinion:}
    \begin{itemize}
        \item \texttt{\{OPINION\}}
    \end{itemize}

    \item \textbf{Familiarize yourself with these key terms and commands in the CCS:}
    \begin{itemize}
        \item \texttt{〈SOL〉}: Start of a 2D closed curve
        \item \texttt{L (Line)}: \(x, y\) - Coordinates of the line's end-point
        \item \texttt{A (Arc)}: \(x, y\) - Coordinates of the arc's end-point; \(\alpha\) - Sweep angle; \(f\) - Counter-clockwise flag
        \item \texttt{R (Circle)}: \(x, y\) - Center coordinates; \(r\) - Radius
        \item \texttt{E (Extrude)}: \(\theta, \phi, \gamma\) - Orientation of the sketch plane; \(p_x, p_y, p_z\) - Origin of the sketch plane; \(s\) - Scale factor; \(e_1, e_2\) - Extrude distances; \(b\) - Boolean operation type; \(u\) - Extrude direction
    \end{itemize}

    \item \textbf{Pay special attention to the Extrude command parameters:}
    \begin{itemize}
        \item \texttt{Parameter \(b\)}: "NewBodyFeatureOperation", "JoinFeatureOperation", "CutFeatureOperation", or "IntersectFeatureOperation"
        \item \texttt{Parameter \(u\)}: "OneSideFeatureExtentType", "SymmetricFeatureExtentType", or "TwoSidesFeatureExtentType"
    \end{itemize}

    \item \textbf{Analyze the differences between the CAD description and the real CCS. }

    \item \textbf{Based on your analysis, modify the CAD parametric description to address any issues found.}

    \item \textbf{Output CAD parametric description ONLY!!!}
\end{enumerate}

\\

\midrule

LLMCAD Enhanced CCS &

You are a CAD sequence generation parameter. Your task is to generate a correct CAD Command Sequence (CCS) based on a CAD description, existing CCS, and confidence levels. Follow these instructions carefully:

Now, consider the following CAD description:  
\texttt{\{CAD\_DESCRIPTION\}}

Here is the existing CCS and its associated confidence levels:  
\texttt{\{EXISTING\_CCS\}}  
\texttt{\{CONFIDENCE\}}

Analyze the confidence levels provided. Pay special attention to parameters with confidence levels below 0.98, as these should be the focus of your modifications. Based on the CAD description and the confidence levels, generate a new CCS. Modify the existing CCS to align with the description, focusing on adjusting parameters with low confidence levels. Ensure that the new CCS accurately represents the described CAD operations.

Output the new CCS without any explanations or additional comments.

\\

\end{xltabular}
\end{onecolumn}

\end{document}